%% file: distributionalmeasures.tex
\documentclass[fullname]{clv2}

\usepackage{amstext}
\usepackage{epsf}
\usepackage{graphicx}
\usepackage{latexsym}
\usepackage{mathpazo}
\usepackage{wasysym}
\usepackage{lscape}
\usepackage{longtable}

\issue{1}{1}{2006}

\dochead{}

\runningtitle{Distributional Measures of Semantic Distance}

\runningauthor{Mohammad and Hirst}

\begin{document}

\title{Distributional Measures of Semantic Distance: A Survey}

\author{Saif Mohammad}
\affil{University of Toronto}

\author{Graeme Hirst}
\affil{University of Toronto}

\maketitle

\setlength\textheight{49.61pc}

\begin{abstract}
The ability to mimic human notions of semantic distance has
widespread applications. Some measures rely only on raw text
(distributional measures) and some rely on knowledge sources such as WordNet.
Although extensive studies have been performed to compare
WordNet-based measures with human judgment, 
 the use of distributional measures
as proxies to estimate semantic distance has received little attention. 
 Even though they have traditionally performed poorly when compared to WordNet-based
measures, they lay claim to certain uniquely attractive features, 
such as their applicability in resource-poor languages and
their ability to mimic both semantic similarity and semantic relatedness. 
Therefore, this paper presents a detailed study of distributional measures.
Particular attention is paid to flesh out 
the strengths and limitations of both WordNet-based and distributional measures,
and how distributional measures of distance can be brought more in line with human notions of semantic distance.
We conclude with a brief discussion of recent work on hybrid measures.
\end{abstract}

\section{Introduction}

{\bf Semantic distance} is a measure of how close or
distant the meanings of two units of language are.
The units of language may be words, phrases, sentences, paragraphs, or
documents. 
The nouns {\it dance}
and {\it choreography}, for example, are closer in meaning than the
nouns {\it clown} and {\it bridge},
and so are said to be semantically closer.
The semantic distances between words (or more precisely, between concepts)
can be used as fundamental building blocks for measuring semantic distance between 
larger units of language. 
The ability to mimic
human judgments of semantic distance is useful in numerous natural language tasks
including machine translation, word sense
disambiguation, thesaurus creation, information retrieval, text
summarization, and identifying discourse structure.
This paper describes the state-of-the-art in corpus-based measures of semantic distance
between these fundamental units of language.
It identifies the significant challenges that existing approaches to semantic
distance face and in the process fleshes out questions that lead to
a better understanding of why two concepts are considered semantically close.
The paper concludes with a discussion of new hybrid approaches, that 
show the potential to address these challenges.

Units of language, especially words, 
may have more than one possible meaning. However, their context may be used
to determine the intended senses. For example, 
{\it star} can mean both {\sc celestial body} and {\sc celebrity}; however, {\it star} 
in---{\it Stars are powered by nuclear fusion}---refers only to {\sc celestial body} and is much closer to {\it sun} than to {\it famous}.
Thus, semantic distance between words in context is in fact the distance
between their underlying senses or lexical concepts. 
Therefore, in this paper, we take word senses to be a
particular kind of concept. When we refer directly to a concept (written in small capitals),
it is with the understanding that it is the sense of one or more words, as reflected in the name that
we give the concept. We do not, in this paper, consider concepts that are unlexicalized.

Humans consider two concepts to be semantically close if there is a sharing of some meaning.
Specifically, two lexical concepts are semantically close if there is a {\bf lexical semantic relation}
between the concepts. Putting it differently, the reason why two concepts are considered semantically close
can be attributed to a lexical semantic relation that binds them. 
According to \namecite{Cruse86}, a lexical semantic relation is a relation between {\bf lexical units}---a surface form along with a sense. 
As he points out, the number of semantic relations that bind concepts is innumerable; but certain
relations, such as hyponymy, meronymy, antonymy, and troponymy, are more systematic and have enjoyed more attention in the linguistics
community. However, as \namecite{MorrisH04} point out, these relations are far out-numbered
by others, which they call {\bf non-classical relations}. Here are a few of the kinds of non-classical
relations they observed: positive qualities ({\sc brilliant, kind}), concepts pertaining
to a concept ({\sc kind, chivalrous, formal} pertaining to {\sc gentlemanly}), and
commonly co-occurring words (locations such as {\sc homeless, shelter}; problem--solution
pairs such as {\sc homeless, shelter}).

\subsection{Semantic relatedness and semantic similarity}
Semantic distance is of two kinds: {\bf semantic similarity} and {\bf semantic relatedness}.
The former is a subset of the latter, 
but the two terms
may be used interchangeably in certain contexts, making it even more important to be
aware of their distinction.
Two concepts are considered to be semantically similar if there is a synonymy (or near-synonymy),
hyponymy (hypernymy), antonymy, or troponymy relation between them (examples include {\sc apples--bananas},
{\sc doctor--surgeon}, {\sc dark--bright}). 
Two word senses are considered to be semantically related if there is any
lexical semantic relation at all between them---classical or non-classical
(examples include {\sc apples--bananas}, {\sc surgeon--scalpel}, {\sc tree--shade}).

\subsection{Human judgments of semantic distance}
Humans are adept at estimating
semantic distance; 
but consider the following questions:
How strongly will two people agree/disagree
on distance estimates? Will the agreement vary over different sets of concepts?
Are we equally good at estimating semantic similarity and semantic relatedness?
In our minds, is there a clear distinction between related and unrelated concepts
or are concept-pairs spread across the whole range from synonymous to unrelated?
Some of the earliest work that begins to
address these questions is by \namecite{RubensteinG65}.
They conducted quantitative experiments with human subjects (51 in all)
who were asked to rate 65 English word pairs on a scale from 0.0 to 4.0 as per 
their semantic distance. 
The word pairs chosen ranged from almost synonymous to unrelated. However, they were all 
noun pairs and those that were semantically close were also semantically similar; the dataset did not 
contain word pairs that are semantically related but not semantically similar.
The subjects repeated the annotation after two weeks and the new distance values
had a Pearson's correlation $r$ of 0.85 with the old ones.
\namecite{MillerC91} also 
conducted a similar study on 30 word pairs taken from the Rubenstein-Goodenough
pairs. These annotations had a high correlation ($r = 0.97$) with the mean annotations
of \namecite{RubensteinG65}.
\namecite{Resnik99} repeated these experiments and found the inter-annotator correlation ($r$)
to be 0.90.

\namecite{ResnikD00} conducted annotations of 48 verb pairs and found 
inter-annotator correlation  ($r$) to be 0.76 (when the verbs were 
presented without context) and 0.79 (when presented in context).
\namecite{Gurevych2005} and \namecite{ZeschGM07} 
 asked native German speakers to mark two different
sets of German word pairs with distance values. Set 1 
was a German translation of the
\namecite{RubensteinG65} dataset. It had 65 noun--noun word
pairs. Set 2 was a larger dataset containing 350 word
pairs made up of nouns, verbs, and adjectives.  The semantically close
word pairs in the 65-word set were mostly synonyms or hypernyms (hyponyms) of each
other, whereas those in the 350-word set had both classical and non-classical
relations with each other.  Details of these {\bf
  semantic distance benchmarks}
are summarized in
Table~\ref{tab:datasets}. Inter-subject correlations (last column in Table \ref{tab:datasets})
are indicative of the degree of ease in annotating the datasets.

\begin{table*}
    \begin{center}
	\caption{Different datasets that are manually annotated with distance values. Pearson's correlation coefficient ($r$) was used to determine inter-annotator correlation (last column).}
    \label{tab:datasets}
	\resizebox{\textwidth}{!}{
        \begin{tabular}{lcccccc}
        \hline
        {\bf Dataset} & {\bf Year} & {\bf Language} & {\bf \# pairs} & {\bf PoS}  & {\bf \# subjects} & {\bf $r$} \\
        \hline
        \hline
        Rubenstein and Goodenough  	& 1965 	& English  	& 65  	& N  	& 51    & -    \\
        Miller and Charles  		& 1991 	& English  	& 30  	& N  	& 38     & .90    \\
        Resnik and Diab  			& 2000 	& English  	& 27  	& V  	& 10     & .76 and .79 \\
        Gurevych  					& 2005 & German  & 65  & N           & 24    & .81    \\
        Zesch and Gurevych 			& 2006 & German  & 350 & N, V, A     & 8     & .69    \\
        \hline
        \end{tabular}
	}
    \end{center}
	Note: Rubenstein and Goodenough (1965) do not report inter-subject correlation, but determine intra-subject correlation to be 0.85 for 36 (out of the 65) word pairs for which similarity judgments were repeated by 15 (of the 51) subjects.
    \normalsize
\end{table*}

It should be noted here that even though the annotators were presented with word-pairs
and not concept-pairs, it is reasonable to assume that they were annotated as per their
closest senses. For example, given the noun pair {\em bank} and {\em interest}, most if not all
will identify it as semantically related even though both words have more than one sense
and many of the sense--sense combinations are unrelated (for example, the {\sc river bank}
sense of {\it bank} and the {\sc special attention} sense of {\it interest}).
The high agreement and correlation values suggest that humans are quite good and consistent at estimating
semantic distance of noun-pairs; however, annotating verbs and adjectives and a 
combination of parts of speech is harder. This also means that estimating
semantic relatedness is harder than estimating semantic similarity.

Apart from showing that humans can indeed estimate semantic distance, these 
datasets act as ``gold standards" to evaluate automatic
distance measures.
However, lack of large amounts of data from human subject experimentation
limits the reliability of this mode of evaluation. Therefore automatic distance
measures are also evaluated by their usefulness in natural language tasks
such as those mentioned earlier.

\subsection{Automatic measures of semantic distance}
Automatic measures of semantic distance quantify the semantic distance between word pairs.
They give values within a certain range (for example, 0 to 1), such that one end of this range
represent maximal closeness or synonymy, while the other end represents maximal distance.
 Depending on which end is which, measures of semantic distance can be classified as
 {\bf measures of distance} (larger values indicate greater distance and less closeness) and 
 {\bf measures of closeness} (larger values indicate shorter distance and more closeness).\footnote{A note
 about terminology: In many contexts, the term {\it distance measures} refers to the complete set of measures
 (irrespective of what the different ends of the range signify). In certain other contexts (as in this paragraph),
 {\it distance measures} refers only to those measures that give larger values to signify greater distance.
 The context, usually by its reference to this numeric property or lack thereof will make clear the intended meaning of the term.}
A measure of closeness can be easily converted to a measure of distance by applying a suitable
inverse function, 
or vice versa. 

Two classes of automatic methods have been traditionally used to determine semantic
distance.  {\bf Knowledge-rich measures of concept-distance}, such as those
of \namecite{JiangC97}, \namecite{LeacockC98}, and \namecite{Resnik95},
rely on the structure of a knowledge
source, such as WordNet, to determine the distance between two
concepts defined in it.\footnote{The nodes in WordNet (synsets) represent word senses 
and edges between nodes represent semantic relations such as hyponymy and meronymy.}
{\bf Distributional measures of word-distance (knowledge-lean measures)}, such as cosine and
$\alpha$-skew divergence~\cite{Lee01}, rely on the {\bf distributional hypothesis},
which states that two words tend to be semantically close 
if they occur in similar contexts \cite{Firth57}.
Distributional measures rely simply on text (and possibly some shallow syntactic processing)
and can give the distance between any two
words that occur at least a few times.

The various WordNet-based measures have been widely studied \cite{BudanitskyH06,PatwardhanBT03}.
The study of distributional measures on the whole has
received much less attention.\footnote{See \namecite{Curran04} and \namecite{WeedsWM04} for other work that compares various distributional 
measures.}
Even though, as \namecite{Weeds03} and \namecite{MohammadH06b} show, they perform poorly when compared to WordNet-based
measures, the distributional measures of word-distance have many attractive features, including their
ability to measure both semantic similarity and semantic relatedness. Further,
they are not dependent on costly knowledge sources that do not exist for most languages.
This paper therefore focuses on distributional measures and analyzes
their strengths and limitations. Particular attention is paid to the different
kinds of distributional measures and their components.
The motivation is that a better understanding of distributional measures
will lead to bringing them more in line with human notions of semantic distance,
while still maintaining their applicability to resource-poor languages and their
ability to mimic both semantic similarity and semantic distance.

\section{Knowledge-rich approaches to semantic distance}
Before we begin our examination of distributional measures, we look
briefly at the resource-based measures. In some ways they are 
complementary to distributional measures and so the discussion will 
set the context for the analysis of distributional measures.

Creation of electronically available ontologies and semantic
networks such as WordNet has allowed their use to help solve numerous
natural language problems including the measurement of semantic
distance.
\namecite{BudanitskyH06}, \namecite{HirstB05}, and \namecite{PatwardhanBT03}
have done an extensive survey of the various WordNet-based measures,
their
comparisons with human judgment on selected word pairs, and
their usefulness in applications such as real-word spelling correction
and word sense disambiguation. Hence, this section provides
only a brief summary of the major knowledge-rich  measures
of semantic distance.

\subsection{Measures that exploit WordNet's semantic network}
\label{s:Wnet}

A number of WordNet-based measures consider two concepts to be close if
they are close to each other in WordNet.
One of the earliest and simplest measures is Rada et al.'s \shortcite{RadaMBB89}
{\bf edge-counting} method. The shortest path in the network
between the two target concepts ({\bf target path}) is determined.
The more edges there are
between two words, the more distant they are. Elegant as it
may be, the measure
hinges on the largely incorrect assumption that all the network edges
correspond to identical semantic distance. 

Nodes in a network may be connected by different kinds of
lexical relations such as hyponymy, meronymy, and so on.
Edge counts apart, Hirst and St-Onge's \shortcite{HirstS98} measure takes into account
the fact that if the target path consists of edges that belong
to many different relations, then the target concepts are likely
more distant. The idea is that if we start from a particular node
$c_1$
and take a path via a particular relation (say, hyponymy),
to a certain extent the concepts reached will be semantically related
to $c_1$. However, if during the way we take edges
belonging to different relations (other than hyponymy), very
soon we may reach words that are unrelated. Hirst and St-Onge's
measure of semantic relatedness is:
\begin{equation}
\text{\itshape HS}(c_1,c_2) = C - \text{\itshape path length} - k \times d
\end{equation}
\noindent where $c_1$ and $c_2$ are the target concepts,
$d$ is the number of times an edge pertaining to a relation different
from that of the preceding edge is taken, and
$C$ and $k$ are empirically determined constants.
More recently, \namecite{YangP05} proposed a weighted edge-counting method to determine
semantic relatedness using the hypernymy/hyponymy, holonymy/meronymy,
and antonymy links in WordNet.

\namecite{LeacockC98}
used just one relation (hyponymy) and modified the path length
formula to reflect the fact that edges lower down in the hyponymy
hierarchy correspond to smaller semantic distance than the ones
higher up. For example, synsets pertaining to {\it sports car} and {\it car} (low in the hierarchy)
are much
more similar than those pertaining to {\it transport} and {\it instrumentation} (higher
up in the hierarchy) even
though both pairs of nodes are separated by exactly one edge in the
hierarchy. Their formula is:
\begin{equation}
{\text{\itshape LC}}(c_1,c_2) =  -\log \frac{{\text{\itshape len}}(c_1,c_2)}{2D}
\end{equation}
\noindent where $D$ is the maximum depth of the taxonomy.

\namecite{Resnik95} suggested a measure that combines corpus statistics
with WordNet. He proposed that since the 
{\bf lowest common subsumer} or {\bf lowest superordinate (lso)}
of the target nodes represents what is similar between them,
the semantic similarity between the two concepts
is directly proportional to how specific the lso is.
The more general the lso is, the larger the semantic distance between the target nodes.
This specificity is measured by the formula for information
content (IC)---the negative logarithm of the probability of the lso:
\begin{equation}
{\text{\itshape Res}}(c_1,c_2) = {\text{\itshape IC}}(lso(c_1,c_2)) = -\log {\text{\itshape P}}(lso(c_1,c_2))
\end{equation}
\noindent Observe that using information content has the effect of inherently
scaling the semantic similarity measure by the depth of the taxonomy. Usually, the lower
the lowest superordinate, the lower the probability of occurrence of
the lso and the concepts subsumed by it, and hence, the higher its
information content.

As per Resnik's formula, given a particular lowest superordinate, the exact positions
of the target nodes below it in the hierarchy do not have any effect
on the semantic similarity. Intuitively, we would expect that word pairs closer
to the lso are more semantically similar than those that are distant. \namecite{JiangC97} and \namecite{Lin97}
incorporate this notion into their measures which
are arithmetic variations of the same terms.
The \namecite{JiangC97} measure ({\itshape JC}\/) determines how dissimilar each target concept
is from the lso ($IC(c_1) - IC(lso(c_1,c_2))$ and $IC(c_2) - IC(lso(c_1,c_2))$). The final semantic
distance between the two concepts is
then taken to be the sum of these differences.
\namecite{Lin97} (like Resnik) points out that the lso is what is common between the two
target concepts and that its information content is the
common information between the two concepts. His formula ({\itshape Lin}\/) can be thought of
as taking the Dice coefficient of the information in the two target concepts.
\vspace*{-4mm}
\begin{eqnarray}
{\text{\itshape JC}}(c_1,c_2)& =&  2\log p({\text{\itshape lso}}(c_1,c_2))
- (\log(p(c_1)) + (\log(p(c_2))) \\
{\text{\itshape Lin}}(c_1,c_2)& =&  \frac{2 \times \log p({\text{\itshape lso}}(c_1,c_2))}
{\log(p(c_1)) + (\log(p(c_2))}
\end{eqnarray}

\vspace*{-4mm}
\namecite{BudanitskyH06} showed that the Jiang-Conrath measure
has the highest correlation (0.850) with the Miller and Charles noun pairs
and performs better than all these measures in a spelling correction task. \namecite{PatwardhanBT03}
achieved similar results using the measure for word sense disambiguation.

All of the approaches described above rely heavily (if not solely) on the
hypernymy/hyponymy network in WordNet; they are designed for, and evaluated on, noun--noun pairs.
However, more recently, \namecite{ResnikD00} and \namecite{YangP06a} developed
measures aimed at verb--verb pairs. \namecite{ResnikD00} ported several measures
which are traditionally applied on the noun hypernymy/hyponymy network
(edge counting, and the measures of \namecite{Resnik95}, and \namecite{Lin97})
to the relatively shallow verb troponymy network. The two information content--based measures
ranked a carefully chosen set of 48 verbs best in order of their semantic distance.\footnote{Only those
verbs were selected which require a theme, and the sub-categorization frames had to match.}
\namecite{YangP06a} ported their earlier work on nouns \cite{YangP05} to verbs.
In order to compensate for the relatively shallow verb troponymy hierarchy
and the lack of a corresponding holonymy/meronymy hierarchy, they proposed
several back-off models---the most useful one being the distance between
a noun pair that has the same lexical form as the verb pair. However, the approach
has too many tuned parameters (9 in all) and performed poorly on a set of 36
TOEFL word-choice questions involving verb targets and alternatives.

\subsection{Measures that rely on dictionaries and thesauri}
 \hspace{0.0pt} \namecite{Lesk86} introduced a method to perform word sense disambiguation
using word glosses (definitions). The glosses of the senses of a target word are
compared with those of its context and the number of word overlaps is determined.
The sense with the greatest number of overlaps is chosen as the intended sense of the target.
Inspired by this approach, \namecite{BanerjeeP03} proposed a semantic relatedness measure
that deems two concepts to be more semantically related if there is more overlap in their
glosses. Notably, they overcome the problem of short glosses by considering the glosses
of concepts related to the target concepts through the WordNet lexical semantic relations
such as hyponymy/hypernymy. They also give more weight to larger overlap sequences.
\namecite{PatwardhanP06} proposed another gloss-based semantic relatedness measure
which performed slightly worse than the extended gloss overlap measure in a word
sense disambiguation task, but markedly better at ranking the \namecite{MillerC91} word pairs.
They create
{\bf aggregate co-occurrence vectors} for a WordNet sense by adding
the co-occurrence vectors of the words in its WordNet gloss.
The distance between two senses is then determined by the
cosine of the angle between their aggregate vectors.
Such aggregate co-occurrence
vectors are expected to be noisy because they are created
from data that is not sense-annotated.

\namecite{JarmaszSzpakowicz2003} use the taxonomic structure of {\it Roget's Thesaurus}
to determine semantic similarity. Two words are considered maximally similar if
they occur in the same semicolon group in the thesaurus. Then on, decreasing in similarity
are word pairs in the same paragraph, words pairs in different paragraphs belonging to the same part of speech and within the same
category, word pairs  in the category, and so on until word pairs which have nothing in common except
that they are in the thesaurus (maximally distant). They show that this simple
approach performs remarkably well at ranking word pairs and determining the 
correct answer in sets of TOEFL, ESL, and {\it Reader's Digest} word choice problems.

\subsection{Challenges}
In this section, we review some of the shortcomings of resource-based
measures in order to motivate and to compare them with distributional measures 
that we will introduce in Section \ref{s:dist}.

\subsubsection{Lack of high-quality WordNet-like knowledge sources}
\label{s:Lbottleneck}

Ontologies, WordNets, and semantic networks are available for a few
languages such as English, German, and Hindi.
Creating them requires human experts and it is time intensive.
Thus, for most languages, we cannot use WordNet-based measures simply due to the
lack of a WordNet in that language. Further,
even if created, updating an ontology is again expensive and there
is usually a lag between the current state of language usage/comprehension
and the semantic network representing it. Further, the complexity
of human languages makes creation of even a near-perfect semantic
network of its concepts impossible. Thus in many ways the
ontology-based measures are only as good as the networks on which
they are based. 

On the other hand, distributional measures require only text.
Large corpora, billions of
words in size, may now be collected by a simple web crawler.
Large corpora of more-formal writing are also available (for
example, the {\em Wall Street Journal} or the {\em American Printing
House for the Blind (APHB)} corpus). This makes distributional
measures very attractive.

\subsubsection{Poor estimation of semantic relatedness}
\label{s:}

As \namecite{MorrisH04} pointed out, a large number
of concept pairs, such as {\sc strawberry--cream} and {\sc doctor--scalpel},
have a non-classical relation between them ({\sc strawberries} are usually eaten with {\sc cream} and
a {\sc doctor} uses a {\sc scalpel} to make an incision).
These words are not semantically similar, but rather semantically related.
An ontology- or WordNet-based measure will correctly identify the amount of
semantic relatedness only if such relations are explicitly coded into the knowledge source.
Further, the most accurate WordNet-based measures rely only on its extensive is-a hierarchy.
This is because networks of other lexical-relations such as meronymy are
much less developed. Further, the networks for different parts of speech
are not well connected.
All this means that, while WordNet-based measures accurately estimate semantic
similarity between nouns, their estimation of semantic relatedness, especially in
pairs other than noun--noun, is at best poor and at worse non-existent.
On the other hand, distributional measures can be used to determine both
semantic relatedness and semantic similarity.

\subsubsection{Inability to cater to specific domains}
\label{s:}

Given a concept pair, measures that rely only on WordNet and no text, such as that of \namecite{RadaMBB89},
give just one distance value.
However, two concepts may be very close in a certain domain but
not so much in another. For example, {\sc space} and {\sc time}
are close in the domain of quantum mechanics but not so much
in most others. Ontologies have been made for specific domains, which
may be used to determine semantic similarity specific to these domains. However,
the number of such ontologies is very limited. Some of the more successful
WordNet-based measures, such as that of \namecite{JiangC97}, that rely also on text,
do indeed capture domain-specificity to some extent, but the distance
values are still largely shaped by the underlying network, which is not domain-specific.
On the other hand, distributional measures rely primarily (if not completely)
on text, and large amounts of
corpora specific to particular domains can easily be collected.

\subsubsection{Computational complexity and storage requirements}
\label{s:Lcomp}

As applications for linguistic distance become more
sophisticated and demanding, it becomes attractive to
pre-compute and store the distance values between all
possible pairs of words or senses.  
However both WordNet-based and distributional 
measures have large space requirements to do this, requiring
matrices of size $N \times N$, where $N$ is very large.
In case of WordNet-based measures, $N$ is the number of senses (81,000 just for nouns). 
In case of distributional measures, $N$ is the size of the
vocabulary (at least 100,000 for most languages).
 Given that the above matrices tend to be sparse\footnote{Even though WordNet-based
 and distributional measures give non-zero closeness values to a large number of
 term pairs, values below a suitable threshold 
 can be reset to 0.} and that computational
 capabilities are continuing to improve, the above limitation may not seem hugely problematic,
 but as we see more and more natural language applications in embedded systems and hand-held devices, such as
 cell phones, iPods, and medical equipment, memory and computational power become serious constraints.

\subsubsection{Reluctance to cross the language barrier}
\label{s:Lcross}

Both WordNet-based and distributional measures have largely been used in
a monolingual framework. Even though semantic distance 
seems to hold promise in tasks such as machine translation and multi-lingual text
summarization that inherently involve two or more languages, automatic measures
of semantic distance have rarely been applied.
With the development of the EuroWordNet, involving interconnected networks
of seven different languages, it is possible that we shall see more cross-lingual
work using WordNet-based measures in the future. However, such an interconnected
network will be very hard to create for more-different
language pairs such as English and Chinese or English and Arabic.

\section{Knowledge-lean, distributional  approaches to semantic distance}
\label{s:dist}

\subsection{The distributional hypotheses: the original and the revised}
\label{s:disthypo}

{\bf Distributional measures} are inspired by the maxim ``You
shall know a word by the company it keeps'' \cite{Firth57}.
These measures rely simply on raw text and possibly some shallow
syntactic processing. They are
much less resource-hungry than the semantic measures, but
they measure the distance between words rather than
word-senses or concepts. Two words are
considered close if they occur in similar contexts.
The context of a target word is usually taken to be the set of words within a certain
window around it, for example, $\pm 5$ words or the complete sentence.
The set of contexts of a target word is usually represented by the set of words 
in these contexts, their strength of association (SoA) with the target word, and 
possibly their syntactic relation with the target, for example verb--object, subject--verb, and so on.
The strength of co-occurrence association between the target and another word quantifies how much more (or less)
than chance the two words occur together in text. Commonly used measures of association
are conditional probability (CP) and pointwise mutual information (PMI).
The distance between the sets of contexts of two target words can be used as a proxy for their semantic distance,
as words found in similar contexts tend to be semantically similar---the 
{\bf distributional hypothesis} \cite{Firth57,Harris68}.

The hypothesis makes intuitive sense, as \namecite{BudanitskyH06} point out:
If two words have many co-occurring
words in common, then similar things are being said about both of them and
so they are likely to be semantically similar. Conversely, if two words are
semantically similar,
then they are likely to be used in a similar fashion in text and thus
end up with many common co-occurrences. For example, the semantically
similar {\em bug} and {\em insect} are expected to have a number of common
co-occurring words such as {\em crawl, squash, small, woods}, and so on,
in a large-enough text corpus.

The distributional hypothesis only mentions semantic similarity and not 
semantic relatedness. 
This, coupled with the fact that the difference between
semantic relatedness and semantic similarity is somewhat nuanced and can
be missed, meant that almost all work employing the distributional hypothesis
was labeled as estimating semantic similarity. However, it should be noted
that distributional measures can be used to estimate both semantic similarity and
semantic relatedness. Even though \namecite{SchutzeP97} and \namecite{LandauerFL98}, for example,
use the term {\it similarity} and not {\it relatedness}, their
LSA-based distance measures in fact estimate semantic relatedness and not semantic
similarity.
We propose more-specific distributional hypotheses that make clear how
distributional measures can be used to estimate semantic similarity
and how they can be used to measure semantic relatedness:
\begin{quote}
{\bf Hypothesis of the distributionally close and semantically related:} \\
Two target words are distributionally close and semantically related if they have
many common strongly co-occurring words. \\
(For example, {\it doctor}--{\it surgeon} and
{\it doctor}--{\it scalpel}. See example co-occurring words in Table \ref{tab:simrel}.)
\end{quote}

\begin{quote}
{\bf Hypothesis of the distributionally close and semantically similar:} \\
Two target words are distributionally close and semantically similar if they have
many common strongly co-occurring words that each have the same
syntactic relation with the two targets. \\
(For example, {\it doctor}--{\it surgeon},
but not {\it doctor}--{\it scalpel}. See syntactic relations with example co-occurring 
words in Table \ref{tab:simrel}.)
\end{quote}

\begin{table}
 \caption[Example: Common syntactic relations of target words with co-occurring words.]
 {Example: Common syntactic relations of target words with co-occurring words.}
         \label{tab:simrel}
         \begin{center}
         \hspace{-0.05in}

\resizebox{\textwidth}{!}{
         \begin{tabular}{lcccccc} \hline

                        	 					& &\multicolumn{5}{c}{\bf Co-occurring words} \\ 
                        			 			& &{\it cut} (v) 		& &{\it hardworking} (adj) 	&  &{\it patient} (n) \\ \hline
                        {\bf Semantically similar} 			    			& &					& &					& & \\ 
                        {\bf target pair} 			& &     								& &					& & \\ 
                        {\it doctor} (n) 			& &subject--verb   	& &noun--qualifier	& &subject--object \\ 
                        {\it surgeon} (n) 			& &subject--verb    & &noun--qualifier	& &subject--object \\ 
                        			 			& &     								& &					& & \\ 
                        {\bf Semantically related} 			    			& &					& &					& & \\ 
                        {\bf target pair} 			& &     			& &									& & \\ 
                        {\it doctor} (n) 			& &subject--verb    & &noun--qualifier	& &subject--object 	 \\ 
                        {\it scalpel} (n) 			& &prepositional object--verb & &--				& &prepositional object--object \\ 
            \hline
         \end{tabular}
}
         \end{center}
         \hspace{-0.20in}
         \end{table}

\noindent The idea is that both semantically similar and semantically related word pairs
will have many common co-occurring words. However, words that are semantically similar
belong to the same broad part of speech (noun, verb, etc.), but the same need not
be true for words that are semantically related.
Therefore, words that are semantically similar will tend to have 
the same syntactic relation, such as verb--object or subject--verb,
with most common co-occurring words. 
Thus, the two words are considered semantically related simply if they
have many common co-occurring words. But to be semantically similar as well,
the words must have the same syntactic relation with co-occurring words.
Consider the word pair {\em doctor--operate}.
In a large enough body of text, the two words are likely to
have the following common co-occurring words: {\em patient, scalpel,
surgery, recuperate}, and so on. All these words
will contribute to a high score of relatedness.
However, they do not have the same syntactic relation with the two targets. (The word
{\em doctor} is almost always used as a noun while {\em operate}
is a verb.) Thus, as per the two revised distributional hypotheses,
{\em doctor} and {\em operate} will correctly be identified as semantically related
but not semantically similar. 
The word pair {\em doctor--nurse},
on the other hand, will be identified as both semantically related and semantically similar.

In order to clearly differentiate from the distance as calculated
by a WordNet-based semantic measure (described earlier in Section \ref{s:Wnet}), the distance calculated by a 
corpus-based distributional measure will be referred to as
{\bf distributional distance}. 

\subsection{Corpus-based measures of distributional distance}
We now describe specific distributional measures that rely on the distributional hypotheses;
depending on which specific hypothesis
they use, they mimic either semantic similarity or semantic relatedness.
	
	\subsubsection{Spatial Metrics: Cos, ${\text L}_1$, ${\text L}_2$}

Consider a multidimensional space in which
the number of dimensions is equal to the size of the vocabulary. A word
$w$ can be represented by a point in this space such that
the component of $\vec w$ in a
dimension (corresponding to word $x$, say) is equal to the strength of association (SoA)
of $w$ with $x$ (${\text \it{SoA}}(w,x)$). 
Thus, the vectors corresponding to two words are {\it close} together,
and thereby get a low distributional distance score, if they share many co-occurring words
and the co-occurring words have more or less the same strength of association
with the two target words.
The distance between two vectors can be calculated in
different ways as described below.

 \paragraph{Cosine}
 \label{s:cosine}

The {\bf cosine} method (denoted by $\text{\bf Cos}$) is one of the earliest and most widely used distributional measures.
Given two words $w_1$ and $w_2$, the cosine measure calculates
the cosine of the angle between $\vec w_1 $ and $\vec w_2 $.
If a large number of words co-occur with both $w_1$ and $w_2$,
then $\vec w_1 $ and $\vec w_2$ will have a small angle between
them and the cosine will be large; signifying a large
relatedness/similarity between them. The cosine measure gives
scores in the range from $0$ (unrelated) to $1$ (synonymous). 
So the higher the value, the less distant the target word-pair is.
\begin{equation}
\label{eq:relCos}
\text{\itshape Cos}(w_1,w_2) = \frac{\sum_{w \in C(w_1) \cup C(w_2)} \left( P(w|w_1) \times P(w|w_2) \right) }
{\sqrt{ \sum_{w \in C(w_1)} P(w|w_1)^2 } \times \sqrt{ \sum_{w \in C(w_2)} P(w|w_2)^2 } }
\end{equation}
\noindent where $C(w)$ is the set of words that co-occur (within a certain window)
with the word $w$ in a corpus.
In this instantiation of the cosine measure, conditional probability
of the co-occurring words given the target words is used as the strength
of association.

The cosine was used, among others, by \namecite{SchutzeP97} and
\namecite{YoshidaYK03}, who suggest methods of automatically generating distributional thesauri from
text corpora. \namecite{SchutzeP97} use the
Tipster category B corpus~\cite{Tipster} (450,000 unique terms)
and the {\em Wall Street Journal} to create a large but sparse
co-occurrence matrix of 3,000 medium-frequency words (frequency
rank between 2,000 and 5,000). Latent semantic indexing
(singular value decomposition) \cite{SchutzeP97} is used to reduce the dimensionality
of the matrix and get for each term a word vector of its 20 strongest
co-occurrences. The cosine of a target word's vector with each of the other
word vectors is calculated and the 
words that give the highest scores comprise the thesaurus entry for the target word.

\namecite{YoshidaYK03} believe
that words that are closely related for one person may be
distant for another. They use
around 40,000 HTML documents to generate personalized thesauri
for six different people. Documents used to create the thesaurus
for a person are retrieved from the subject's home page and a web crawler which
accesses linked documents. The authors also suggest a root-mean-squared
method to determine the similarity of two different thesaurus
entries for the same word.

\paragraph{Manhattan and Euclidean Distances}

Distance between two points (words) in vector space can also
be calculated using the formulae for {\bf Manhattan distance} a.k.a.\@ the ${\bf L_1}$ {\bf norm}
(denoted by ${\bf L_1}$)
or {\bf Euclidean distance} a.k.a.\@ the {\bf ${\text L}_2$ norm} (denoted by ${\text L}_2$).
In the Manhattan distance~(\ref{eq:L1}) (\namecite{DaganLP97}, \namecite{DaganLP99},
and \namecite{Lee99}), the difference in strength of association of
$w_1$ and $w_2$ with each word that they co-occur with is summed.
The greater the difference, the greater is the distributional distance between the two words.
Euclidean distance~(\ref{eq:L2}) \cite{Lee99} employs the root mean square of the
difference in association to get the final distributional distance.
Both the ${\text L}_1$ and ${\text L}_2$ norms give scores in the range between 0 (zero
distance or synonymous) and infinity (maximally distant or unrelated).

\begin{eqnarray}
\label{eq:L1}
L_1(w_1,w_2)& =& \sum_{w \in C(w_1) \cup C(w_2)} \mid P(w|w_1) - P(w|w_2) \mid \\
\label{eq:L2}
L_2(w_1,w_2)& =& \sqrt{\sum_{w \in C(w_1) \cup C(w_2)} \left(P\left(w|w_1\right) - P\left(w|w_2\right)\right)^2 }
\end{eqnarray}

\noindent The above formulae use conditional probability
of the co-occurring words given a target word as the strength
of association.

\namecite{Lee99} compared the ability of all three spatial metrics to
determine the probability of an unseen (not found in training data) word
pair. The measures in order of their performance (from better to worse) were: ${\text L}_1$ norm,
cosine, and ${\text L}_2$ norm.
\namecite{Weeds03} determined the correlation of word pair ranking as per
a handful of distributional measures with human rankings (\namecite{MillerC91} word pairs). 
She used verb-object pairs from the {\em British
National Corpus (BNC)} and
found the correlation of ${\text L}_1$ norm with human rankings to be 0.39.

\subsubsection{Mutual information--based measures: Hindle, Lin}

 \hspace{0.0pt} \namecite{Hindle90} was one of the first to factor the strength
of association of co-occurring words into a distributional similarity measure.\footnote{See \namecite{Grefenstette92}
for an approach that does not incorporate strength of association of co-occurring words. He, like \namecite{Hindle90}, uses syntactic dependencies to
to characterize the set of contexts of a target word. The Jaccard coefficient is used to determine how similar the two sets of contexts are.}
Consider the nouns $n_j$ and
$n_k$ that exist as objects of verb $v_i$ in different
instances within a text corpus. Hindle used the following formula
to determine the distributional similarity of $n_j$ and $n_k$
solely from their occurrences as object of $v_i$:
\begin{equation}
\label{eq:Hindle1}
\text{\itshape Hin}_{\text{\itshape obj}} (v_i, n_j, n_k) = \left\{ \begin{array}{l} \min(I(v_i,n_j), I(v_i,n_k)),\\ \qquad \quad \text{if}\; I(v_i,n_j) > 0\; \text{and}\; I(v_i,n_k) > 0 \\
                                                 \mid \max(I(v_i, n_j), I(v_i, n_k))\mid,\\ \qquad \quad   \text{if}\; I(v_i, n_j) < 0\; \text{and}\; I(v_i, n_k) < 0 \\
                                                 0, \quad \quad \text{otherwise} \end{array} \right.
\end{equation}
\noindent $I(n, v)$ stands for the pointwise mutual information (PMI)
between the noun $n$ and verb $v$
(note that in case of negative PMI values,
the maximum function captures the PMI which is lower in absolute value).
The measure follows from the distributional hypothesis---the more similar the associations of co-occurring words
with the two target words, the more semantically similar they are.
Hindle used PMI\footnote{\namecite{Hindle90} and
\namecite{Lin98C} both refer to pointwise mutual information as mutual
information.}  as the strength of association.
Using the minimum of
the two PMIs captures the similarity in the strength of association of $v_i$
with each of the two nouns. 

Hindle used an analogous formula to calculate
distributional similarity ($Hin_{subj}$) using the subject--verb relation. The overall
distributional similarity between any two nouns is calculated by the formula:
\begin{equation}
\label{eq:Hindle2}
\text{\itshape Hin}(n_1,n_2) = \sum_{i = 0}^{N} \left( \text{\itshape Hin}_{\text {\itshape obj}}(v_i, n_1, n_2) + \text{\itshape Hin}_{\text{\itshape subj}}(v_i, n_1, n_2) \right)
\end{equation}
\noindent The measure
gives similarity scores from 0 (maximally dissimilar) to infinity (maximally similar or synonymous).
Note that in Hindle's measure, the set of co-occurring words used is restricted
to include only those words that have the same syntactic relation with both
target words (either verb--object or verb--subject). This is 
therefore a measure
that mimics semantic similarity and not semantic relatedness. A form of
Hindle's measure where all co-occurring words are used, making it a measure 
that mimics semantic relatedness, is shown below:
\begin{equation}
\label{eq:Hindle3}
\text{\itshape Hin}_{\text{\itshape{rel}}}(w_1,w_2) = \sum_{w \in C(w)} \left\{ \begin{array}{l} \min(I(w,w_1), I(w,w_2)),\\ \qquad \quad \text{if}\; I(w,w_1) > 0\; \text{and}\; I(w,w_2) > 0 \\
                                                 \mid\max(I(w, w_1), I(w, w_2))\mid,\\ \qquad \quad   \text{if}\; I(w, w_1) < 0\; \text{and}\; I(w, w_2) < 0 \\
                                                 0, \quad \quad \text{otherwise} \end{array} \right.
\end{equation}
\noindent where $C(w)$ is the set of words that co-occur with word $w$.

\namecite{Lin98C} suggests a different measure derived from his
information-theoretic definition of similarity \cite{Lin98B}. Further,
he uses a broad set of syntactic relations apart from just subject--verb and
verb--object relations and shows that using multiple  relations is beneficial even for
Hindle's measure. He first extracts triples of the form $(x,r,y)$ from
the partially parsed text, where the word $x$ is related to $y$ by the
syntactic relation $r$. 
Lin defines the distributional similarity between two words, $w_1$ and $w_2$, as follows:
\begin{equation}
\label{eq:LinCorpus}
\text{\itshape Lin}(w_1,w_2) = \frac{\sum_{(r,w)\, \in\, T(w_{1})\, \cap\, T(w_{2})} \left(I(w_{1},r,w) + I(w_{2},r,w)\right)}
           {{\sum_{(r,w')\, \in\, T(w_1)} I(w_1,r,w') + \sum_{(r,w'')\, \in\, T(w_2)} I(w_2,r,w'')}}
\end{equation}
\noindent where $T(x)$ is the set of all word pairs $(r,y)$ such that the pointwise mutual information
$I(x,r,y)$, is positive. Note that this is different from \namecite{Hindle90}
where even the cases of negative PMI were considered. 
\namecite{ChurchH89} showed that it is hard to accurately predict
negative word association ratios with confidence, and so, co-occurrence pairs with negative
PMI are ignored. The measure gives similarity scores from 0 (maximally dissimilar) to 1 (maximally similar).

Like Hindle's measure, Lin's
is a measure of distributional {\it similarity}. However,
it distinguishes itself from that of Hindle in two respects.
First, Lin normalizes the similarity score between two words (numerator of (\ref{eq:LinCorpus}))
by their cumulative strengths of association with the rest of the co-occurring words (denominator of (\ref{eq:LinCorpus})). 
This is a
significant improvement as now high PMI of the target words
with shared co-occurring words alone does not guarantee a high distributional similarity score.
As an additional requirement, the target words must have low PMI
with words they do not both co-occur with.
Second, Hindle uses the minimum of
the PMI between each of the target words and the shared
co-occurring word, while Lin uses the sum. Taking
the sum has the drawback of not penalizing for a mismatch in strength
of co-occurrence, as long as $w_1$ and $w_2$ both co-occur with a word.

\namecite{Hindle90} used a portion of the {\em Associated Press} news stories
(6 million words) to classify the nouns into semantically related
classes. \namecite{Lin98C} used his measure to generate a distributional thesaurus
from a 64-million-word corpus of the {\em Wall Street Journal, San Jose Mercury},
and {\em AP Newswire}. He also provides a
framework for evaluating such automatically generated thesauri by comparing
them with WordNet-based and Roget-based thesauri. He shows that the distributional thesaurus
created with his measure is closer to the WordNet and Roget-based thesauri than
that created using Hindle's measure.

\subsubsection{Relative Entropy--Based Measures: KLD, ASD, JSD}

\paragraph{Kullback-Leibler divergence}

Given two probability mass functions
$p(x)$ and $q(x)$, their {\bf relative entropy} $D(p\Vert q)$ is:

\begin{equation}
D(p\Vert q) = \sum_{x \in X} p(x) \log \frac{p(x)}{q(x)} \hspace{1in} \text {for } q(x) \ne 0
\end{equation}

\noindent Intuitively, if $p(x)$ is the accurate probability mass function corresponding
to a random variable $X$, then $D(p\Vert q)$ is the information lost when
approximating $p(x)$ by $q(x)$. In other words, $D(p\Vert q)$ is
indicative of how different the two distributions are. Relative
entropy is also called the {\bf Kullback-Leibler divergence} or the
{\bf Kullback-Leibler distance} (denoted by {\bf KLD}).

\namecite{PereiraTL93} and \namecite{DaganPL94} point
out that words have probabilistic distributions with respect to neighboring
syntactically related words. For example, there exists a certain probabilistic
distribution ($d_1 (P(v|n_1))$, say) of a particular noun $n_1$ being the object of any verb.
This distribution can be estimated by corpus counts of parsed or chunked  text.
Let $d_2$ ($P(v|n_2)$) be the corresponding distribution for noun $n_2$.
These distributions ($d_1$ and $d_2$) define the contexts of the two nouns ($n_1$
and $n_2$, respectively). As per the distributional hypothesis,
the more these contexts are similar, the more $n_1$ and $n_2$ are semantically similar.
Thus the Kullback-Leibler distance between the two distributions
is indicative of the semantic distance between the nouns $n_1$ and $n_2$.
\begin{equation}
\begin{array}{rcll}
\text{\itshape KLD}(n_1,n_2)& =& D(d_1\Vert d_2) &  \\
                            & =& \sum_{v \in \text {\it Vb}} P(v|n_1) \log \frac{P(v|n_1)}{P(v|n_2)} & \text {for } P(v|n_2) \ne 0 \\
                            & =& \sum_{v \in \text {\it Vb}^{\prime} (n_1) \cap \text {\it Vb}^{\prime} (n_2)} P(v|n_1) \log \frac{P(v|n_1)}{P(v|n_2)} & \text {for } P(v|n_2) \ne 0
\end{array}
\end{equation}
\noindent where $\text {\itshape Vb}$ is the set of all verbs and $\text{\itshape Vb}^{\prime} (x)$
is the set of verbs that have $x$ as the object.
Note again that the set of co-occurring words used is restricted
to include only verbs that each have the same syntactic relation (verb--object) with both
target nouns. This too is therefore a 
measure that mimics semantic similarity and not semantic relatedness.

It should be noted that the verb--object relationship is not inherent to the
measure and that one or more of any other syntactic relations may be used.
One may also estimate semantic relatedness by using all words
co-occurring with the target words.
Thus a more generic expression of the
Kullback-Leibler divergence is as follows:
\begin{equation}
\begin{array}{rcll}
\label{eq:KLD}
\text{\itshape KLD}(w_1,w_2)& =& D(d_1\Vert d_2) & \\
                            & =& \sum_{w \in V} P(w|w_1) \log \frac{P(w|w_1)}{P(w|w_2)} & \text {for } P(w|w_2) \ne 0 \\
                            & =& \sum_{w \in C(w_1) \cup C(w_2)} P(w|w_1) \log \frac{P(w|w_1)}{P(w|w_2)} & \text {for } P(w|w_2) \ne 0
\end{array}
\end{equation}
\noindent where $V$ is the vocabulary (all the words found in a corpus).
$C(w)$, as mentioned earlier, is the set of words occurring (within
a certain window) with word $w$. 

It should be noted that the Kullback-Leibler distance is not symmetric; that
is, the distance from $w_1$ to $w_2$ is not necessarily, and even not likely,
the same as the distance from $w_2$ to $w_1$. This asymmetry is
counterintuitive to the general notion of semantic similarity of words, although
\namecite{Weeds03} has argued in favor of asymmetric measures.
Further, it is very likely that there are instances such that $P(w_1|v)$
is greater than 0 for a particular verb $v$, while due to data sparseness
or grammatical and semantic constraints,
the training data has no sentence where $v$ has the object $w_2$. This makes
$P(w_2|v)$ equal to 0 and the ratio of the two probabilities
infinite. Kullback-Leibler divergence is not defined in such cases, but approximations
may be made by considering smoothed values for the denominator.

\namecite{PereiraTL93} used KLD to create
clusters of nouns from verb-object pairs corresponding
to the thousand most frequent nouns in the {\itshape Grolier's Encyclopedia}, June 1991
version (10 million words).
\namecite{DaganPL94} used KLD to estimate
the probabilities of bigrams that were not seen in a text corpus. They point
out that a significant number of possible bigrams are not seen in any
given text corpus. The probabilities of such bigrams may be determined
by taking a weighted average of the probabilities of bigrams composed
of distributionally similar words. Use of Kullback-Leibler distance as the semantic
distance metric yielded a 20\% improvement in perplexity on the {\em Wall
Street Journal} and dictation corpora provided by ARPA's HLT program~\cite{Paul91}.

It should be noted here that the use of distributionally similar words
to estimate unseen bigram probabilities will likely lead to erroneous results
in case of less-preferred and strongly-preferred collocations (word pairs).
\namecite{DianaH02} point out that even though words like {\em task} and {\em job}
are semantically very similar, the collocations they form with other words
may have varying degrees of usage. While {\em daunting task} is a
strongly-preferred collocation, {\em daunting job} is rarely used. Thus
using the probability of one bigram to estimate that of another will
not be beneficial in such cases.\\

\paragraph{$\alpha$-skew divergence}

The {\bf $\alpha$-skew divergence} ({\em ASD}) is a slight modification of the Kullback-Leibler
divergence that obviates the need for smoothed probabilities. It has the
following formula:
\begin{equation}
\label{eq:alpha}
\text{\itshape ASD}(w_1,w_2) = \sum_{w \in C(w_1) \cup C(w_2)} P(w|w_1) \log \frac{P(w|w_1)}{\alpha  P(w|w_2) + (1 - \alpha) P(w|w_1)}
\end{equation}
\noindent where $\alpha$ is a parameter that may be varied but is usually set to $0.99$.
Note that the denominator within the logarithm is never zero with a non-zero
numerator. Also, the measure retains the asymmetric nature of the
Kullback-Leibler divergence.
\namecite{Lee01} shows that $\alpha$-skew divergence performs better than
Kullback-Leibler divergence in estimating word co-occurrence probabilities. \namecite{Weeds03}
achieves a correlation of $0.48$ and $0.26$ with human judgment on the Miller and Charles
word pairs using $ASD(w_1,w_2)$ and $ASD(w_2,w_1)$, respectively.\\

\paragraph{Jensen-Shannon divergence}

A relative entropy--based measure that overcomes the problem of asymmetry in
Kullback-Leibler divergence is the {\bf Jensen-Shannon divergence}
a.k.a.\@ {\bf total divergence to the average} a.k.a.\@ {\bf information radius}.
It is denoted by {\bf JSD} and has the following formula:
\begin{eqnarray}
\label{eq:JSD}
\text{\itshape JSD}(w_1,w_2)& =& D\left(d_1 \Vert \frac{1}{2}(d_1 + d_2)\right) + D\left(d_2 \Vert \frac{1}{2}(d_1 + d_2)\right) \\
& =& \sum_{w \in C(w_1) \cup C(w_2)} \Bigg( P(w|w_1) \log \frac{P(w|w_1)}{\frac{1}{2}\left(P(w|w_1) + P(w|w_2)\right)} + \nonumber\\
& & \qquad \qquad P(w|w_2) \log \frac{P(w|w_2)}{\frac{1}{2}\left(P(w|w_1) + P(w|w_2)\right)} \Bigg)
\end{eqnarray}
\noindent The Jensen-Shannon divergence is the sum of the Kullback-Leibler divergence between each of the individual
co-occurrence distributions $d_1$ and $d_2$ of the target words with the average distribution ($\frac{d_1 + d_2}{2}$).
Further, it can be shown that the Jensen-Shannon divergence avoids the problem
of zero denominator. The Jensen-Shannon divergence
is therefore always well defined and, like $\alpha$-skew divergence, obviates the need for smoothed estimates.

The Kullback-Leibler divergence, $\alpha$-skew divergence, and Jensen-Shannon divergence
all give distributional distance scores from 0 (synonymous) to infinity (unrelated).

\subsubsection{Latent Semantic Analysis}

\hspace{0.0pt} \namecite{LandauerFL98} proposed {\bf Latent semantic analysis (LSA)}, which can be used to determine distributional
distance between words or between sets of words.\footnote{\namecite{LandauerFL98}
describe it as a measure of {\em similarity}, but in fact it is a distributional
measure that mimics semantic relatedness.}
Unlike the various approaches described earlier where a word--word co-occurrence
matrix is created, the first step of LSA involves the creation of a word--paragraph,
word--document, or similar such word-passage matrix, where a {\it passage} is some grouping of words.
A cell for word $w$ and passage $p$ is populated with the number of times $w$
occurs in $p$ or, for even better results, a function of this frequency that captures how much information
the occurrence of the word in a text passage carries. 

Next, the dimensionality
of this matrix is reduced by applying {\bf singular value decomposition (SVD)}, a 
standard matrix decomposition technique. This smaller set of dimensions represents
abstract (unknown) concepts. Then the original word--passage matrix is recreated,
but this time from the reduced dimensions. 
\namecite{LandauerFL98} point out that this results in new matrix cell values that
are different from what they were before. 
More specifically, words that are expected to occur more often in a passage than
what the original cell values reflect are incremented.
Then a standard vector distance measure, such as cosine, that captures the 
distance between distributions of the two target words is applied. 

LSA was used by \namecite{SchutzeP97,Turney2001} and \namecite{Rapp03}
to measure distributional distance, with encouraging results.
However, there is no non-heuristic way to determine when the
dimension reduction should stop. Further, the generic concepts represented by the
reduced dimensions are not interpretable; that is, one cannot determine
which concepts they represent in a given sense inventory. This means
that LSA cannot directly be used for tasks such
as unsupervised sense disambiguation or estimating semantic similarity of known concepts.
 LSA is computationally expensive as singular value decomposition,
 a key component for dimensionality reduction, requires computationally intensive matrix operations.
 This makes LSA less scalable to large amounts of text \cite{GormanC06}.
Finally, it too, like other distributional word-distance 
measures conflates the many senses of a word
(see Section \ref{s:Lconflation} ahead for more discussion on sense conflation).

\subsubsection{Co-occurrence Retrieval Models}

The distributional measures suggested by \namecite{Weeds03} are based on a notion of
substitutability: the more appropriate it is to substitute word $w_1$ 
in place of word $w_2$ in a suitable natural language task, 
the more semantically similar they are.  
She uses {\bf co-occurrence retrieval} (the retrieval of words that co-occur with a target word
from text) to determine the degree to which one word is substitutable by another.
The degree of substitutability
of $w_2$ with $w_1$ is dependent on how the proportion of co-occurrences of $w_1$ 
that are also co-occurrences of $w_2$ and the proportion of co-occurrences of $w_2$ 
that are also co-occurrences of $w_1$. 
Thus Weeds's distributional
measures have a precision component and a recall component (which may or may not incorporate
the strength of co-occurrence association). The final
score is a weighted sum of the precision, recall, and standard $F$ measure
(see equation~(\ref{eq:CRMfinal})\footnote{$P$ is
short for $P(w_1,w_2)$, while $R$ is short for $R(w_1,w_2)$. The abbreviations
are made due to space constraints and to improve readability.}).
The weights determine the importance of precision and recall and are
determined empirically.
If precision and recall are equally important, then it results in a symmetric measure which gives 
identical scores for the distributional similarity of $w_1$ with $w_2$ and 
$w_2$ with $w_1$. Otherwise, we get an asymmetric measure which
assigns different scores to the two cases. 
\begin{equation}
\label{eq:CRMfinal}
CRM(w_1,w_2) = \gamma \Biggl[ \frac{2 \times P \times R}{P + R} \Biggr]  +  (1 - \gamma) \Biggl[ \beta [ P ] + (1 - \beta) [R] \Biggr]
\end{equation}
\noindent $\gamma$ and $\beta$ are tuned parameters that lie between 0 and 1.

Both precision and recall can be considered as the product of a 
core formula (denoted by $core$) and a penalty function (denoted by $penalty$).
{Weeds03} provides six (three times two) distinct formulae for precision
and recall, depending on the strength of co-occurrence (three alternatives)
and whether or not a penalty is applied for differences in strength of association
of common co-occurring words (two alternatives).

Depending on the strength of association, the CRMs are classified as 
{\bf type-based, token-based,} and {\bf mutual information--based}.
The CRMs that use simple counts of the common co-occurrences
and not the strength of associations
as core precision and recall values are called type-based
CRMs (denoted by the superscript {\em type}). 
The CRMs that use conditional probabilities of the shared co-occurring words
with the target words are called token-based
CRMs (denoted by the superscript {\em token}). 
The CRMs that use pointwise mutual information of the shared co-occurring words with target words
are called mutual information--based
CRMs (denoted by the superscript {\em mi}). The core precision
and recall formulae for type, token, and mutual information--based
CRMs are listed below:

\begin{eqnarray}
 \text{core}_P^{type\ {}\ {}}(w_1,w_2)&     =& \frac{\mid C(w_1) \cap C(w_2) \mid}{\mid C(w_1) \mid} \\
 \text{core}_R^{type\ {}\ {}}(w_1,w_2)&        =& \frac{\mid C(w_1) \cap C(w_2) \mid}{\mid C(w_2) \mid} \\
 \text{core}_P^{token\/}(w_1,w_2)&    =& \sum_{w \in C(w_1) \cap C(w_2)} P(w|w_1) \\
 \text{core}_R^{token\/}(w_1,w_2)&       =& \sum_{w \in C(w_1) \cap C(w_2)} P(w|w_2) \\
 \text{core}_P^{mi\ {}\ {}\ {}}(w_1,w_2)&       =& \frac{\sum_{w \in C(w_1) \cap C(w_2)} I(w,w_1)}{\sum_{w \in C(w_1)} I(w,w_1)} \\
 \text{core}_R^{mi\ {}\ {}\ {}}(w_1,w_2)&          =& \frac{\sum_{w \in C(w_1) \cap C(w_2)} I(w,w_2)}{\sum_{w \in C(w_2)} I(w,w_2)} 
\end{eqnarray}

\noindent where $C(x)$ is the set of words that co-occur with $x$. 

Depending on the penalty function, the CRMs are classified as 
{\bf additive} and {\bf difference-weighted}.
The CRMs that do not
penalize difference in strength of co-occurrence are called
additive CRMs (denoted by the subscript {\em add});
those that do penalize are called difference-weighted CRMs
(subscript {\em dw}). The penalty is 
a conditional probability--based function (\ref{eq:penalty P type}, \ref{eq:penalty R type}) for the token- and type-based
CRMs, and a mutual information--based function (\ref{eq:penalty P mi}, \ref{eq:penalty R mi}) for the mutual information--based
CRM. 
\begin{eqnarray}
\label{eq:penalty P type}
penalty_{P}^{type} = penalty_{P}^{token}& =& \frac{\min(P(w|w_1),P(w|w_2))}{P(w|w_1)} \\
\label{eq:penalty R type}
penalty_{R}^{type} = penalty_{R}^{token}& =& \frac{\min(P(w|w_1),P(w|w_2))}{P(w|w_2)} \\
\label{eq:penalty P mi}
penalty_{P}^{mi}& = &\frac{\min(I(w,w_1),I(w,w_2))}{I(w,w_1)} \\
\label{eq:penalty R mi}
penalty_{R}^{mi}& =& \frac{\min(I(w,w_1),I(w,w_2))}{I(w,w_2)} 
\end{eqnarray}
The  six pairs of precision and recall difference-weighted CRMs are thus as follows: 
\begin{eqnarray}
& P_{add}^{type}(w_1,w_2) =& \frac{\mid C(w_1) \cap C(w_2) \mid}{\mid C(w_1) \mid} \\
& R_{add}^{type}(w_1,w_2) =& \frac{\mid C(w_1) \cap C(w_2) \mid}{\mid C(w_2) \mid} \\
& P_{dw}^{type}(w_1,w_2) =& \frac{\sum_{\mid C(w_1) \cap C(w_2) \mid} \frac{\min(P(w|w_1),P(w|w_2))}{P(w|w_1)}}{\mid C(w_1) \mid} \\
& R_{dw}^{type}(w_1,w_2) =& \frac{\sum_{\mid C(w_1) \cap C(w_2) \mid} \frac{\min(P(w|w_1),P(w|w_2))}{P(w|w_2)}}{\mid C(w_2) \mid}
\end{eqnarray}

\begin{eqnarray}
P_{add}^{token}(w_1,w_2)& =& \sum_{w \in C(w_1) \cap C(w_2)} P(w|w_1) \\
R_{add}^{token}(w_1,w_2)& =& \sum_{w \in C(w_1) \cap C(w_2)} P(w|w_2) \\
P_{dw}^{token}(w_1,w_2)& =& \sum_{w \in C(w_1) \cap C(w_2)} \min(P(w|w_1),P(w|w_2)) \\
R_{dw}^{token}(w_1,w_2)& =& \sum_{w \in C(w_1) \cap C(w_2)} \min(P(w|w_2),P(w|w_1))
\end{eqnarray}

\begin{eqnarray}
P_{add}^{mi}(w_1,w_2)& =& \frac{\sum_{w \in C(w_1) \cap C(w_2)} I(w,w_1)}{\sum_{w \in C(w_1)} I(w,w_1)} \\
R_{add}^{mi}(w_1,w_2)& =& \frac{\sum_{w \in C(w_1) \cap C(w_2)} I(w,w_2)}{\sum_{w \in C(w_2)} I(w,w_2)} \\
P_{dw}^{mi}(w_1,w_2)& =& \frac{\sum_{w \in C(w_1) \cap C(w_2)} \min(I(w,w_1),I(w,w_2))}{\sum_{w \in C(w_1)} I(w,w_1)} \\
R_{dw}^{mi}(w_1,w_2)& =& \frac{\sum_{w \in C(w_1) \cap C(w_2)} \min(I(w,w_1),I(w,w_2))}{\sum_{w \in C(w_2)} I(w,w_2)}
\end{eqnarray}

\noindent Note that in case of the difference-weighted token and mutual information--based precision
and recall formulae, there is a cancellation of a pair of terms obtained
from the core formulae and the penalty.

Asymmetry in substitutability is intuitive,
as in many cases
it may be acceptable to substitute a word, say {\em dog}, with another, say {\em animal}, but the
reverse is not acceptable as often. Since Weeds uses
substitutability as a measure of semantic similarity, she believes that
distributional similarity between two words should reflect this property
as well. Hence, like the Kullback-Leibler divergence, all her 
distributional similarity models are asymmetric. 

\namecite{Weeds03} extracted verb--object pairs of 2,000 nouns from the {\em British
National Corpus (BNC)}. The verbs related to the target words
by the verb--object relation were used. Thus each of the co-occurring
verbs is related to the target nouns by the same syntactic relation
and therefore the measures mimic semantic similarity,
not relatedness.
Correlation with human judgment (Miller and Charles word pairs) showed that 
difference-weighted ($r=0.61$) and additive mutual information--based measures
($r=0.62$) performed far better than the other CRMs.

\section{The anatomy of a distributional measure}
\label{s:anatomy}

Even though there are numerous distributional measures, many of which may seem
dramatically different from each other,
all distributional measures
perform two functions: (1) create {\bf distributional profiles (DPs)}, and (2) calculate
the distance between two DPs.

The distributional profile of a word is the strength of association between it and
each of the lexical, syntactic, and/or semantic units that
co-occur with it. Commonly used {\bf measures of strength of
association} are conditional probability (0 to 1) and
pointwise mutual information ($-\infty$ to $\infty$).
Commonly used units of co-occurrence with the target
are other {\it words}, and so we speak of the {\bf lexical
distributional profile of a word (lexical DPW)}.   
The
co-occurring words may be all those in a predetermined
window around the target, or may be restricted to those that
have a certain syntactic ({\it e.g.,} verb--object) or
semantic ({\it e.g.,} agent--theme) relation with the target
word.  We will refer to the former kind of DPs as {\bf relation-free}.
Usually in the latter
case, separate association values are calculated for each of
the different relations between the target and the
co-occurring units. We will refer to such DPs as {\bf
relation-constrained}.
\begin{table}
 \caption[Measures of DP distance, measures of strength of association, and standard combinations.]
 {Measures of DP distance, measures of strength of association, and standard combinations.
  Measures of strength of association that are traditionally used are marked in bold. The use of
  other measures of association remains to be explored.
 }
         \label{tab:distrib}
         \begin{center}
         \hspace{-0.05in}

         \begin{tabular}{lll} \cline{1-1} \cline{3-3} 

                        {\bf Measures of DP distance} 		&$\;\;\;\;\;\;\;\;$ &{\bf Measures of strength of association}\\ \cline{1-1} \cline{3-3} 
                        $\alpha$-skew divergence (ASD)   	& &$\phi$ coefficient (Phi)				 \\
                        cosine (Cos)                     	& &{\bf conditional probability (CP)}	 \\ 
						Dice coefficient (Dice)			& &cosine (Cos)									 \\
						Euclidean distance (${\text L}_2$ norm)		& &Dice coefficient (Dice)			 \\
						Hindle's measure (Hin)				& &odds ratio (Odds)					     \\
						Kullback-Leibler divergence	(KLD)	& &{\bf pointwise mutual information (PMI)}	 \\
						Manhattan distance	(${\text L}_1$ norm)	& &Yule's coefficient (Yule) 		 \\  
            			Jensen--Shannon divergence (JSD)	& &										 \\
                        Lin's measure (Lin)              	& &										 \\
            \cline{1-1}  \cline{3-3}
         \end{tabular}
         \end{center}
         \hspace{-0.20in}
          \begin{center}
         \begin{tabular}{l} \cline{1-1} 

                        {\bf Standard combinations}\\ \cline{1-1} 
                        $\alpha$-skew divergence with $\phi$ coefficient (ASD--CP)\\
                        cosine with conditional probability (Cos--CP)\\ 
						Dice coefficient with conditional probability (Dice--CP)\\
						Euclidean distance with conditional probability (${\text L}_2$ norm--CP)\\
						Hindle's measure with pointwise mutual information (Hin--PMI)\\
						Kullback-Leibler divergence with conditional probability (KLD--CP)\\
						Manhattan distance with conditional probability (${\text L}_1$ norm--CP)\\ 
            			Jensen--Shannon divergence with conditional probability (JSD--CP)\\
                        Lin's measure with pointwise mutual information (Lin--PMI)\\
            \cline{1-1}  
         \end{tabular}
         \end{center}
         \hspace{-0.20in}
         \end{table}
Typical relation-free DPs are those of \namecite{SchutzeP97} and \namecite{YoshidaYK03}.
Typical relation-constrained DPs are those of
\namecite{Lin98B} and \namecite{Lee01}.
Below are contrived, but plausible, examples of
each for the word {\em pulse};
the numbers are conditional probabilities:
\begin{quote}
{\bf relation-free DP}\\
{\bf \em pulse}: {\em beat} .28, {\em racing} .2, {\em grow} .13,
{\em beans} .09, {\em heart} .04, \ldots
\end{quote}
\begin{quote}
{\bf relation-constrained DP}\\
{\bf \em pulse}: $\langle${\em beat}, subject--verb$\rangle$ .34,
$\langle${\em racing}, noun--qualifying adjective$\rangle$ .22,  $\langle${\em
grow}, subject--verb$\rangle$ .14, \ldots
\end{quote}

Since the DPs represent the contexts of the two target words, the 
distance between the DPs is the distributional distance and, as per the
distributional hypothesis, a proxy for semantic distance.
A {\bf measure of DP distance}, such as cosine, calculates
the distance between two distributional profiles. 
While any of the measures of DP distance may be used with
any of the measures of strength of association, 
in practice only certain combinations are used (see Table~\ref{tab:distrib})
and certain other combinations might not be meaningful (for example,
Kullback-Leibler divergence with $\phi$ coefficient).
Observe from Table~\ref{tab:distrib} that all standard-combination distributional measures
(or at least those that are described in this paper) use either conditional 
probability or PMI as the measure of association.

\subsection{Simple co-occurrences versus syntactically related words}
\hspace{-1mm}   
\namecite{Harris68}, one of the early proponents of the distributional hypothesis,
used syntactically related words to represent the context of a word.
However, the strength of association of any word appearing in the 
context of the target words may be used to determine their distributional similarity.
\namecite{DaganLP97}, \namecite{Lee99}, and \namecite{Weeds03}
represent the context of a noun with verbs whose object it is (single 
syntactic relation),
\namecite{Hindle90} represents the context of a noun with verbs with which 
it shares the verb-object or subject-verb relation, while
\namecite{Lin98C} uses words related to a noun by any of the many pre-decided syntactic relations
to determine distributional similarity.
\namecite{SchutzeP97} and \namecite{YoshidaYK03} use all co-occurring words in a
pre-decided window size.
Although \namecite{Lin98C} shows that the use of multiple syntactic
relations is more beneficial as compared to just one,
\namecite{McCarthyKWC07} show that results obtained using just word co-occurrences
produced almost as good results as those obtained 
using syntactically
related words.
Further, use of syntactically related words entails the
requirement of chunking or parsing the data. 

\subsection{Compositionality}

The various measures of distributional similarity may be divided into two
kinds as per their composition. In certain  measures, each co-occurring 
word contributes to some {\em finite calculable} distributional distance between
the target words. 
The final score of distributional distance is
the sum of these contributions. 
We will call such measures {\bf compositional measures}. 
The relative entropy--based measures,
$L_1$ norm and $L_2$ norm, fall in this category. 
On the other hand,
the cosine measure, along with
Hindle's and Lin's mutual information--based measures,
belong to the category of what we call {\bf non-compositional} measures.
Each co-occurring word shared by both target
words contributes a score to the numerator and the denominator of the measures' formula. 
Words that co-occur with just one of the two target words
contribute scores only to the denominator. The ratio is calculated once all
co-occurring words are considered.
Thus the distributional distance contributed
by individual co-occurrences  is not calculable and the final semantic
distance cannot be broken down into compositional distances contributed
by each of the co-occurrences. 
It is not clear as to which of the two kinds of measures 
(compositional or non-compositional) 
resembles human judgment more closely and how much they differ
in their ranking of word pairs.

\subsubsection{Primary Compositional Measures}

The compositional measures of distributional similarity (or relatedness) capture the contribution
to distance between the target words ($w_1$ and $w_2$) due to a co-occurring word by three
primary mathematical manipulations of the co-occurrence distributions ($d_1$ and $d_2$):
the {\bf difference}, denoted by $\text{\itshape Dif}$ (as in $L_1$ norm), {\bf division}, 
denoted by $\text{\itshape Div}$ (as in the relative entropy--based 
measures), and {\bf product}, denoted by $\text{\itshape Pdt}$ (as in the conditional probability or mutual information--based cosine method). 
We will call the three types of compositional measures {\bf primary compositional
measures (PCM)}. Their form is depicted below:

\begin{eqnarray}
\label{eq:diff}
{\text{\itshape Dif}}& =& \sum_{w \in C(w_1) \cup C(w_2)} \left| P(w|w_1) - P(w|w_2) \right| \\
\label{eq:div}
{\text{\itshape Div}}& =& \sum_{w \in C(w_1) \cup C(w_2)} \left| \log \frac{P(w|w_1)}{P(w|w_2)} \right| \\
\label{eq:pdt}
{\text{\itshape Pdt}}& =& \sum_{w \in C(w_1) \cup C(w_2)} \frac{P(w|w_1) \times P(w|w_2)}{\text{\itshape Scaling Factor}} 
\end{eqnarray}

\noindent Observe that by taking absolute values in expressions (\ref{eq:diff}) and
(\ref{eq:div}), the variation in the distributions for different co-occurring words
has an additive affect rather than one of cancellation. This corresponds to our
distributional hypothesis --- the more the disparity in distributions, the more is
the semantic distance between the target words. The product form (\ref{eq:pdt}) also
achieves this and is based on this theorem:
The product of any two numbers will
always be less than or equal to the square of their average. 
In other words,
the more two numbers are close to each other in value, the higher is the ratio of 
their product to a suitable
scaling factor (for example, the square of their average). 
Note that the difference and division measures 
give higher values when there is large disparity between the strength 
of association of co-occurring words with the target words. They are therefore
measures of distributional distance and not distributional similarity.
The product method gives higher values when the strengths of association
are closer, and is a measure of distributional relatedness.

Although all three methods seem intuitive, each produces different distributional similarity
values and more importantly, given a set of word pairs, each is likely
to rank them differently. For example, consider the division and difference
expressions applied to word pairs ($w_1$, $w_2$) and ($w_3$, $w_4$). For simplicity, let there be
just one word $w'$ in the context of all the words. Given:

\begin{eqnarray*}
P(w'|w_1) = 0.91 \\
P(w'|w_2) = 0.80 \\ 
P(w'|w_3) = 0.60 \\ 
P(w'|w_4) = 0.50 
\end{eqnarray*}

\noindent The distributional distance between word pairs as per the difference PCM:

\begin{eqnarray*}
\text{\itshape Dif\/}(w_1, w_2)& =& | 0.91 - 0.8 | = 0.11 \\
\text{\itshape Dif\/}(w_3, w_4)& =& | 0.6 - 0.5 |  = 0.1 
\end{eqnarray*}

\noindent The distributional distance between word pairs as per the division PCM:

\begin{eqnarray*}
\text{\itshape Div}(w_1, w_2)& =& \left| \log \frac{0.91}{0.8} \right| \; = 0.056 \\
\text{\itshape Div}(w_3, w_4)& =& \left| \log \frac{0.6}{0.5} \right| \; = 0.079 
\end{eqnarray*}

\noindent Observe that for the same set of co-occurrence probabilities, the difference-based
measure ranks the ($w_3, w_4$) pair more distributionally similar (lower distributional distance), while the division-based measure
gives lower distributional similarity values for word pairs having large
co-occurrence probabilities. This behavior is not intuitive and it remains
to be seen, by experimentation, as to which of the three, difference, division
or product, yields distributional similarity measures closest to human notions
of semantic similarity.

The $L_1$ norm is a basic implementation of the difference
method. A simple product-based measure of distributional similarity is as proposed below:

\begin{equation}
\label{eq:prod}
{\text{\itshape Pdt}}^{\text{\itshape Avg}}(w_1,w_2) = \sum_{w \in C(w_1) \cup C(w_2)} \frac{P(w|w_1) \times P(w|w_2)}{(\frac{1}{2}(P(w|w_1) + P(w|w_2)))^2} 
\end{equation}

\noindent The scaling factor used is the square of the average probability.
It can be proved that if the sum of two variables is equal to a constant ($k$, say),
their values must be equal to $k/2$ in order to get the largest product.
Now, let $k$ be equal to the sum of $P(w|w_1)/(P(w|w_1) + P(w|w_2))$ and 
$P(w|w_2)/(P(w|w_1) + P(w|w_2))$. This sum will always be equal to $1$ and hence
the product ($Z$) will be largest only when the two numbers are equal i.e. $P(w|w_1)$
is equal to $P(w|w_2)$. In other words, the farther $P(w|w_1)$ and $P(w|w_2)$
are from their average, the smaller is the product $Z$. Therefore, the measure
gives high scores for low disparity in strengths of co-occurrence and low
scores otherwise.
The incorporation of $1/2$ in the scaling factor results in a measure that ranges between $0$ and
$1$. 

The relative entropy--based methods use a weighted division method. 
Observe that
both Kullback-Leibler divergence  (formula repeated below for convenience --- equation (\ref{eq:KLDII}))
and Jensen-Shannon divergence do not take absolute
values of the division of co-occurrence probabilities. This will mean that
if $P(w|w_1) > P(w|w_2)$, 
the logarithm of their ratio will be positive
and if $P(w|w_1) < P(w|w_2)$, the logarithm will be a negative number.
Therefore, there will be a cancellation of contributions to distributional distance by 
words that have higher co-occurrence probability with respect to $w_1$
and words that have a higher co-occurrence probability with respect to $w_2$.
Observe however that the weight $P(w|w_1)$ multiplying the logarithm means that in general
the positive logarithm values receive higher weight than the negative ones, resulting
in a net positive score. Therefore, with no absolute value of the logarithm, as in the KLD,
the weight plays a crucial role.
A modified Kullback-Leibler divergence ($D^{\text{\itshape Abs}}$) which incorporates 
the absolute value is suggested in equation (\ref{eq:KLDAbs}):\footnote{It should be noted 
that any changes to the formula for Kullback-Leibler divergence means that the resulting 
measure is no longer Kullback-Leibler divergence; these measures are denoted by KLD (and
a suitable subscript and/or superscript simply to indicate that they are derived from the 
Kullback-Leibler divergence.}

\begin{eqnarray}
\label{eq:KLDII}
& &{\text{\itshape KLD}}(w_1,w_2) = D(d_1\Vert d_2) = \sum_{w \in C(w_1) \cup C(w_2)} P(w|w_1) \log \frac{P(w|w_1)}{P(w|w_2)}\\
\label{eq:KLDAbs}
& &{\text{\itshape KLD}}^{\text{\itshape Abs}}(w_1,w_2) = D^{\text{\itshape Abs}}(d_1\Vert d_2) = \sum_{w \in C(w_1) \cup C(w_2)} P(w|w_1) \left| \log \frac{P(w|w_1)}{P(w|w_2)} \right|\nonumber\\
& &
\end{eqnarray}

\noindent The updated Jensen-Shannon divergence measure will remain the same as in 
equation (\ref{eq:JSD}), except that it is a manipulation of $D^{\text{\itshape Abs}}$ 
and not the original Kullback-Leibler divergence (relative entropy).

\begin{equation}
\label{eq:IRadAb}
{\text{\itshape JSD}}^{\text{\itshape Abs}}(w_1,w_2) = D^{\text{\itshape Abs}}(d_1 \Vert \frac{1}{2}(d_1 + d_2)) + D^{\text{\itshape Abs}}(d_2 \Vert \frac{1}{2}(d_1 + d_2))
\end{equation}

\noindent Note that once the absolute value of the logarithm is taken, it no longer makes
much sense to use an asymmetric weight ($P(w|w_1)$) as in the KLD or as necessary to use
a weight at all. Equation~(\ref{eq:UnwKLD}) 
shows a simple
division-based measure. It is an
unweighted form of ${\text{\itshape KLD}}^{\text{\itshape Abs}}(w_1,w_2)$ and so we will call it 
${\text{\em KLD}}_{\text {\em Unw}}^{\text{\itshape Abs}}$.

\begin{eqnarray}
\label{eq:UnwKLD}
\text{\itshape KLD}_{\text{\itshape Unw}}^{\text{\itshape Abs}}(w_1,w_2) = \text{\itshape Div}(w_1,w_2) 
                                                   = \sum_{w \in C(w_1) \cup C(w_2)} \left| \log \frac{P(w|w_1)}{P(w|w_2)} \right| 
\end{eqnarray}

\noindent Experimental evaluation of these suggested modifications of Kullback-Leibler divergence will be interesting.

\subsubsection{Weighting the PCMs}

The performance of the primary compositional measures may be improved
by adding suitable weights to the distributional distance contributed by each co-occurrence.
The idea is that some co-occurrences may be better indicators of semantic
distance than others. Usually, a formulation of the strength of association of the co-occurring
word with the target words is used as weight, the hypothesis being that
a strong co-occurrence is likely to be a strong indicator of semantic closeness.

Weighting the primary compositional measures results in some of the existing
measures. For example, as pointed out earlier, the Kullback-Leibler divergence
is a weighted form of the division measure (not considering the absolute
value). Here, the conditional probability of a co-occurring word with respect
to the first word ($P(w|w_1)$) is used as the weight. Since the absolute value of the logarithm is not
taken and because the weight ($P(w|w_1)$) is
dependent on the first word and not the other, Kullback-Leibler divergence is asymmetric. 
Below is a symmetric weight function:
\begin{eqnarray}
{\text{\itshape weight}}_{\text{\itshape AvgWt}}(w_1,w_2)& =& \frac{1}{2}\left(P(w|w_1) + P(w|w_2)\right)
\end{eqnarray}
\noindent $L_2$ norm is a weighted version of the $L_1$ norm,
the weight being $P(w|w_1) - P(w|w_2)$. 
A simple product measure with 
weights is shown below:
\begin{eqnarray}
\text{\em Pdt}_{\text{\em AvgWt}}^{\text{\em Avg}}& =& \sum_{w \in C(w_1) \cup C(w_2)} \frac{1}{2}(P(w|w_1) + P(w|w_2))
                      \frac{P(w|w_1) \times P(w|w_2)}{(\frac{1}{2}(P(w|w_1) + P(w|w_2)))^2} \nonumber\\
\label{eq:prodWtd}
 & =& \sum_{w \in C(w_1) \cup C(w_2)} \frac{P(w|w_1) \times P(w|w_2)}{\frac{1}{2}(P(w|w_1) + P(w|w_2))}
\end{eqnarray}

A possibly better weight function (which is also symmetric) hinges on  
the following hypothesis: The stronger the association of a co-occurring word with a target word,
the better the indicator of semantic properties of the target word it is.
Equation~(\ref{eq:SaifWtdDiv}) 
shows the corresponding weight function:
\begin{eqnarray}
\label{eq:SaifWtdDiv}
{\text{\itshape weight}}_{\text{\itshape MaxWt}}(w_1,w_2)
 & =& \frac{\max \left(P(w|w_1),P(w|w_2)\right)}
               {\sum_{w' \in C(w_1) \cup C(w_2)}  \max \left(P(w'|w_1),P(w'|w_2)\right)}
\end{eqnarray}
\noindent The co-occurring word is likely to have different 
strengths of associations with the two target words. Taking the maximum
of the two as the weight (\namecite{DaganMM95}) will mean that more weight is
given to a co-occurring word if it has high strength of association with
any of the two target words. As \namecite{DaganMM95} point out, there is strong evidence for dissimilarity 
if the strength of association with the other target word is much lower than the maximum,
and strong evidence of similarity if the strength of association with both
target words is more or less the same. 

\subsection{Predictors of Semantic Relatedness}
Given a pair of target words, the vocabulary may be divided into 
three sets: (1) the set of words that co-occur with both target
words (common); (2) words that co-occur with exactly one of the two target
words (exclusive); (3) words that do not co-occur with either of the two
target words. 
~\namecite{Hindle90} 
uses evidence only from words that co-occur with both target words
to determine the distributional similarity. All the other measures discussed in this paper
so far also use words that co-occur with just one target word.

One can argue that the more there are common co-occurrences between
two words, the more they are related. For example, {\em drink} and 
{\em sip} may be considered related as they have a number of common 
co-occurrences such as {\em water, tea} and so on. Similarly,
{\em drink} and {\em chess} can be deemed unrelated as words that
co-occur with one do not with the other. For example, {\em water}
and {\em tea} do not usually co-occur with {\em chess}, while {\em 
castle} and {\em move} are not found close to {\em drink}. Measures
that use all co-occurrences (common and exclusive) tap into
this intuitive notion.
However, certain strong exclusive co-occurrences can adversely
effect the measure. 
Consider the classic {\em strong coffee}
vs {\em powerful coffee} example (\namecite{Halliday66}). The words {\em
strong} and {\em powerful} are semantically very related.
However, the word {\em coffee} is likely to 
co-occur with {\em strong} but not with {\em powerful}. Further,
{\em strong} and {\em coffee} can be expected to have a large
value of association as given by a suitable measure, say PMI.
This large PMI value, if used in the distributional relatedness formula, can 
greatly reduce the final value. Thus it is not clear if
the benefit of using all co-occurrences is outweighed by the
drawback pointed out.

A further advantage of using only common co-occurrences is that
the Kullback-Leibler divergence can now be used without the
need of smoothed probabilities. 

\begin{equation}
\text{\em KLD}_{\text{\em Com}}(w_1,w_2) = \sum_{w \in C(w_1) \cap C(w_2)} P(w|w_1) \log \frac{P(w|w_1)}{P(w|w_2)}
\end{equation}

\noindent Observe that we are taking the intersection of the 
set of co-occurring words instead of union as in the original
formula (\ref{eq:KLD}).

\subsection{Capitalizing on asymmetry}
Given a hypernym-hyponym pair ({\em automobile-car}, say)
asymmetric distributional measures such as the Kullback-Leibler divergence,
$\alpha$ skew divergence, and the CRMs generate different 
values as the distributional distance of $w_1$ with $w_2$ as compared to that of $w_2$
with $w_1$. Usually, if $w_1$ is a more generic concept than $w_2$,
the measures find $w_1$ to be more distributionally similar to $w_2$ than
the other way round (see \cite{MirkinDG07} for work on lexical entailment using
the Kullback-Leibler divergence). \namecite{Weeds03} argues that this behavior is intuitive as
it is more often acceptable to substitute a generic concept in place of a specific one
than vice versa, and substitutability is a indicator of semantic similarity.

On the other hand, in most cases the notion of asymmetric semantic similarity
is counterintuitive, and possibly detrimental. In many natural language tasks,
one needs the distance between two words and there is no order information.
Further, in case 
two words share a hypernym-hyponym relation, they are likely to be
highly semantically similar. 
Thus given two words, it may make sense to always choose the 
higher of the two distributional similarity values suggested by an asymmetric measure
as the final distributional similarity between the two. This way an asymmetric measure (${\text{\em Sim}}_{\text{\em Asym}}$)
can easily be converted into a symmetric one (${\text{\em Sim}}_{\text{\em Max}}$), 
while still capitalizing 
on the asymmetry to generate more suitable distributional distance values for 
hypernym-hyponym word pairs. Equation~(\ref{eq:SymMax}) states the formula for the
proposed conversion. A specific implementation of the KL divergence
formula is given in equation (\ref{eq:KLDMax}):
\begin{eqnarray}
\label{eq:SymMax}
\text{\em Sim}_{\text{\em Max}}(w_1, w_2)& =& \max (Sim_{Asym}(w_1,w_2), Sim_{Asym}(w_2,w_1)) \\
\label{eq:KLDMax}
\text{\em KLD}_{\text{\em Max}}(w_1, w_2)& =& \max (\text{\em KLD}(w_1,w_2), \text{\em KLD}(w_2,w_1))
\end{eqnarray}
\noindent Another way to convert an asymmetric measure of distributional
distance into a symmetric one is by taking the average (formula~\ref{eq:SymAvg})
of the two possible similarity values. A specific implementation on the KL divergence
formula is given in equations (\ref{eq:KLDAvg}) through (\ref{eq:KLDAvgEnd}):
\begin{eqnarray}
\label{eq:SymAvg}
& &\; \; \text{\em Sim}_{\text{\em Avg}}(w_1, w_2) = \frac{1}{2} (Sim_{Asym}(w_1,w_2) + Sim_{Asym}(w_2,w_1)) \\
\label{eq:KLDAvg}
& &\text{\em KLD}_{\text{\em Avg}}(w_1, w_2) = \frac{1}{2} (\text{\em KLD}(w_1,w_2) + \text{\em KLD}(w_2,w_1)) \\
& &                                         = \frac{1}{2} \sum_{w \in C(w_1) \cup C(w_2)} \left(P(w|w_1) \log \frac{P(w|w_1)}{P(w|w_2)} + P(w|w_2) \log \frac{P(w|w_2)}{P(w|w_1)}\right) \\
& &                                         = \frac{1}{2} \sum_{w \in C(w_1) \cup C(w_2)} \left(P(w|w_1) \log \frac{P(w|w_1)}{P(w|w_2)} - P(w|w_2) \log \frac{P(w|w_1)}{P(w|w_2)} \right)\\
\label{eq:KLDAvgEnd}
& &                                         = \frac{1}{2} \sum_{w \in C(w_1) \cup C(w_2)} \left(P(w|w_1) - P(w|w_2)\right) \log \frac{P(w|w_1)}{P(w|w_2)}
\end{eqnarray}

\subsection{Summarizing the distributional measures}
Table~\ref{tab:distributional measures of distance} 
summarizes the properties of various distributional measures discussed in this paper.

\begin{landscape}
\begin{longtable}{l c c c c c c} \hline \\
         \caption[Measures of distributional distance]{Distributional measures and their properties.}
         \label{tab:distributional measures of distance}\\
            {\bf distributional} &{\bf measure}                   &{\bf compo-}					&      &      &{\bf symm-}                   &{\bf strength of}\\
           {\bf measure}    &{\bf type}		&{\bf sitional}    &{\bf PCM}    		&{\bf formula}     	&{\bf etric}    &{\bf association}\\  \hline \hline
		   \endfirsthead
		\multicolumn{7}{l}{Note: For measures that are not compositional, the type of PCM is not applicable.}
		\endfoot
		   \hline\\
		   \multicolumn{7}{c}{Distributional measures and their properties (continued).}\\
			{\bf distributional} &{\bf measure}                   &{\bf compo-}					&      &               &{\bf symm-}                   &{\bf strength of}\\
           {\bf measure}    &{\bf type}		&{\bf sitional}    &{\bf PCM}    		&{\bf formula}     	&{\bf etric}    &{\bf association}\\ \hline \hline
		   \endhead

			{\em ASD}       &distance        &$\checked$     &division                                        &$\sum_{w \in C(w_1) \cup C(w_2)} P(w|w_1) \log \frac{P(w|w_1)}{\alpha  P(w|w_2) + (1 - \alpha) P(w|w_1)}$      &X      &CP\\
		                    &                &                                                   &                                                                                        &    & &   \\

           $\cos$           &closeness		&X              &n.a. 
                                           &$\frac{\sum_{w \in C(w_1) \cup C(w_2)} \left( P(w|w_1) \times P(w|w_2) \right) }
{\sqrt{ \sum_{w \in C(w_1)} P(w|w_1)^2 } \times \sqrt{ \sum_{w \in C(w_2)} P(w|w_2)^2 } }$     &$\checked$         &CP\\
		                    &                &                                                    &                                                             &                           &    &    \\
           {\em CRMs}       &closeness     &X              &n.a.                                           &$\gamma \biggl[ \frac{2 \times P \times R}{P + R} \biggr]  +  (1 - \gamma) \biggl[ \beta [ P ] + (1 - \beta) [R] \biggr]$    &X        &both\\
                            &               &                                                    &                                                &             &      & \\ 
			${\text{\itshape Dice}}^{\text{\itshape CP}}$    &closeness                                &X     &n.a.                          &$\frac{2 \times \sum_{w \in C(w_1) \cup C(w_2)} \min (P(w|w_1),P(w|w_2))}{ \sum_{w \in C(w_1)} P(w|w_1) + \sum_{w \in C(w_2)} P(w|w_2)} $      &$\checked$         &CP\\ 
                            &               &                                                   &                                                &              &     & \\

           ${\text{\itshape Dif}}\;\; \text{or}\;\; L_1 $   &distance                                 &$\checked$     &difference     &$ \sum_{w \in C(w_1) \cup C(w_2)} \mid P(w|w_1) - P(w|w_2) \mid $                      &$\checked$         &CP\\
                                    &               &       &                                     &  &   &    \\
			${\text{\itshape Div}}$        &distance 			&$\checked$     &division            &$\sum_{w \in C(w_1) \cup C(w_2)} \left| \log \frac{P(w|w_1)}{P(w|w_2)}\right|$                                                      &$\checked$       &CP\\
                                    &                &       &                                 &     &    &    \\ 
		           ${\text{\itshape Hindle}}$   &closeness            &X              &n.a.                                           &$\sum_{w \in C(w)} \left\{ \begin{array}{l} \min(I(w,w_1), I(w,w_2)),\\ \qquad \quad \text{if}\; I(w,w_1) > 0\; \text{and}\; I(w,w_2) > 0 \\
						 \mid \max(I(w, w_1), I(w, w_2))\mid,\\ \qquad \quad   \text{if}\; I(w, w_1) < 0\; \text{and}\; I(w, w_2) < 0 \\
						 0,\\ \qquad \quad \text{otherwise} \end{array} \right.$                                 &$\checked$         &PMI\\

		                    &               &                                                   &                                                               &                         &    &    \\
			${\text{\itshape Jaccard}}^{\text{\itshape CP}}$   &closeness                              &X     &n.a.                                              &$\frac{\sum_{w \in C(w_1) \cup C(w_2)} \min (P(w|w_1),P(w|w_2))}{ \sum_{w \in C(w_1) \cap C(w_2)} \max (P(w|w_1),P(w|w_2))}$
			&$\checked$         &CP\\ 
		                    &               &                                                   &                                                               &                         &    &    \\\hline
           {\em JSD}        &distance                         &$\checked$     &division                                         &$\sum_{w \in C(w_1) \cup C(w_2)} \Bigl( P(w|w_1) \log \frac{P(w|w_1)}{\frac{1}{2}\left(P(w|w_1) + P(w|w_2)\right)} +$
                     &$\checked$         &CP\\
                            &            &    &                                                   &$\qquad \qquad P(w|w_2) \log \frac{P(w|w_2)}{\frac{1}{2}\left(P(w|w_1) + P(w|w_2)\right)} \Bigr)$                 &   &                    \\ 

                                    &               &       &                      &               &    &    \\
	
           $\text{\em KLD}$         &distance                 &$\checked$     &div.                           &$\sum_{w \in C(w_1) \cup C(w_2)} P(w|w_1) \log \frac{P(w|w_1)}{P(w|w_2)}$                                             &X                  &CP\\
                                    &                &       &                                 &     &    &    \\
           $\text{\em KLD}_{\text{\em Com}}$     &distance                     &$\checked$     &div.          &$\sum_{w \in C(w_1) \cap C(w_2)} P(w|w_1) \log \frac{P(w|w_1)}{P(w|w_2)}$                                             &X                  &CP\\
                                    &                &       &                                 &     &    &    \\
           $\text{\em KLD}^{\text{\em Abs}}$     &distance    &$\checked$     &div.                            &$\sum_{w \in C(w_1) \cup C(w_2)} P(w|w_1) \left| \log \frac{P(w|w_1)}{P(w|w_2)} \right|$                  &X                  &CP\\
                                    &                &       &                                 &     &    &    \\

           $\text{\em KLD}_{\text{\em Avg}}$    &distance     &$\checked$     &div.                             &$\frac{1}{2} \sum_{w \in C(w_1) \cup C(w_2)} \left(P(w|w_1) - P(w|w_2)\right) \log \frac{P(w|w_1)}{P(w|w_2)} $    &$\checked$         &CP\\
		                    &                &                                                   &                                                                                        &    & &   \\
           $\text{\em KLD}_{\text{\em Max}}$    &distance     &$\checked$     &div.                           &$\max (\text{\em KLD}(w_1,w_2), \text{\em KLD}(w_2,w_1))$                                                              &$\checked$         &CP\\
		                    &                &                                                   &                                                                                        &    & &   \\

           $L_2$                    &distance                &$\checked$     &difference                          &$\sqrt{\sum_{w \in C(w_1) \cup C(w_2)} \left(P\left(w|w_1\right) - P\left(w|w_2\right)\right)^2 }$       &$\checked$         &CP\\
                                    &                &       &                                 &     &    &    \\
           {\em Lin}                &closeness		&X              &n.a.                                           &$\frac{\sum_{(r,w)\, \in\, T(w_{1})\, \cap\, T(w_{2})} (I(w_{1},r,w) + I(w_{2},r,w))}
            {{\sum_{(r,w')\, \in\, T(w_1)} I(w_1,r,w') + \sum_{(r,w'')\, \in\, T(w_2)} I(w_2,r,w'')}}$              &$\checked$         &PMI\\
                            &               &                                                    &                                                &             &      & \\

           $\text{\itshape{Pdt}}^{\text{\itshape{Avg}}}$     &closeness   &$\checked$     &pdt.                                               &$\sum_{w \in C(w_1) \cup C(w_2)} \frac{P(w|w_1) \times P(w|w_2)}{(\frac{1}{2}(P(w|w_1) + P(w|w_2)))^2}$ 	 	&$\checked$         &CP\\
		                    &               &                                                   &                                                               &                         &    &    \\
           ${\text{\itshape Pdt}}_{\text{\itshape{AvgWt}}}^{\text{\itshape Avg}}$   &closeness  &$\checked$     &pdt.        &$\sum_{w \in C(w_1) \cup C(w_2)} \frac{P(w|w_1) \times P(w|w_2)}{\frac{1}{2}(P(w|w_1) + P(w|w_2))}$      &$\checked$         &CP\\ 
                            &               &                                                    &                                                &             &      & \\ 

							\hline
        \end{longtable}
\end{landscape}

\subsection{Challenges}

\subsubsection{Conflation of word senses}
\label{s:Lconflation}

The distributional hypothesis \cite{Firth57} states that
words that occur in similar contexts tend to be semantically
close. 
But when words have more
than one sense, it is not at all clear what semantic
distance between them actually means.  Further, a word in each of its
senses is likely to co-occur with different sets of words. 
For example, {\em bank} in the {\sc financial institution} sense
is likely to co-occur with {\em interest, money, accounts,}
and so on, whereas the {\sc river bank} sense might have words
such as {\em river, erosion,} and {\em silt} around it.  
Since words that occur together in text tend to refer
to senses that are closest in meaning to one another, in most natural language applications,
what is needed is the distance between the closest senses of the two target words.
However, because distributional measures calculate distance from
occurrences of the target word in all its occurrences and hence all its senses, 
they fail to get the desired result. 
Also note that the dimensionality reduction inherent to latent semantic analysis,
a special kind of distributional measure, has the effect of making the
predominant senses of the words more dominant while de-emphasizing the other senses.
Therefore, an LSA-based
approach will also conflate information from the different senses, and even more
emphasis will be placed on the predominant senses.
Given the semantically close target nouns {\em play} and {\em actor}, for example, a distributional
measure will give a score that is some sort of a dominance-based average of the distances
between their senses.
The noun {\em play} has the predominant sense of {\sc children's recreation} (and not {\sc drama}),
so a distributional measure will tend to give the target pair a large (and thus erroneous) distance score.
WordNet-based measures do not suffer from this problem as they give distance between concepts, not words.

\subsubsection{Lack of explicitly-encoded world knowledge and data sparseness}
\label{s:Lsparse}

It is becoming increasingly clear that more-accurate results can be achieved in 
a large number of natural language tasks, including the estimation of semantic distance, 
by combining corpus statistics with a knowledge source, such as a dictionary, published
thesaurus, or WordNet. 
This is because such knowledge sources capture semantic
information about concepts and, to some extent, world knowledge. 
For example, WordNet, as discussed earlier, has an extensive is-a hierarchy.
If it lists one concept, say {\sc German Shepherd} as a hyponym of another, say {\sc dog}, 
then we can be sure that the two are semantically close. On the other hand, 
distributional measures do not have access to such explicitly encoded information.
Further, unless the corpus used by a distributional measure has sufficient instances of
{\sc German Shepherd} and {\sc dog}, it will be unable to deem them semantically close.
Since Zipf's law seems to hold even for the largest of corpora, there will always be
words that occur too few times to accurately determine their distributional distance from others.

\subsubsection{Limitations shared with WordNet-based measures}
In addition to the limitations described above, which are unique to the knowledge-lean
distributional measures, like the knowledge-rich measures they also suffer from
problems of requiring the calculation of large distance matrices (as described in Section
\ref{s:Lcomp} earlier) and the reluctance to cross the language barrier (Section
\ref{s:Lcross}).

\section{A hybrid approach: Distributional measures of concept-distance}
So far we have looked at approaches that exploit the
structure of a resource such as WordNet, and corpus-based distributional
approaches that make use of co-occurrence statistics. A {\bf hybrid approach to semantic distance}
is one that reconciling the two, combining the information about concepts, explicitly encoded in a linguistic resource,
with the information about words, implicitly encoded in text by co-occurrence.
\namecite{MohammadH06b} and \namecite{MohammadGHZ07} have  proposed one such approach that
combines corpus statistics with a published thesaurus to give the semantic distance between
concepts (rather than words).

\subsection{The distributional hypothesis for concepts}
\label{s:dhc}
 As pointed out in Section \ref{s:Lconflation},  words when used in different senses tend to keep different ``company" (co-occurring words). 
For example, consider the contrived but plausible distributional profile 
of {\it star}: 
\begin{quote}
{\bf \em star$\,$}: {\it space} 0.21, {\it movie} 0.16, {\it famous} 0.15, {\it light} 0.12, {\it constellation} 0.11, {\it heat} 0.08, {\it rich} 0.07, {\it hydrogen} 0.07,~\ldots
\end{quote}
Observe that it has words that co-occur both with {\it star}'s
{\sc celestial body} sense and {\it star}'s {\sc celebrity} sense.
Thus, it is clear that different senses of a word may have very different
distributional profiles. Using a
single DP for the word will mean the union of those
profiles.  While this might be useful for certain
applications, \namecite{MohammadH06b} argue that in a number of tasks
(including estimating semantic distance), acquiring
different DPs for the different senses is not only more
intuitive, but also, as they show through experiments,
more useful.  
They show that the closer the distributional profiles of two concepts, the smaller is their
semantic distance. Below are example distributional profiles of two senses of {\it star}:
\begin{quote}
{\bf {\sc celestial body}}: 
{\it space} 0.36, {\it light} 0.27, {\it constellation} 0.11, {\it hydrogen} 0.07,~\ldots\\
{\bf \sc celebrity}: 
{\it famous} 0.24, {\it movie} 0.14, {\it rich} 0.14, {\it fan} 0.10,~\ldots
\end{quote}
\noindent 
The values are the strength of association (usually pointwise mutual
information or conditional probability) of the target concept
with co-occurring words. 

We have seen that creating distributional profiles of words involves
 simple word--word co-occurrence counts. 
 The creation of DPCs, on the other hand, requires: (1) a concept inventory that list all the concepts and words
 that refer to them, and (2) counts of how often a concept
co-occurs with a word in text. These two aspects will be discussed in the next two sub-sections; however
once created, any of the many distributional measures can be used to estimate the distance
between the DPs of two target concepts (just as in the case of traditional word-distance measures, where
distributional measures are used to estimate the distance between the DPs of two target words).
For example, here is how Mohammad and Hirst adapt the formula for cosine (described earlier in Section \ref{s:cosine})
to estimate distributional distance between two concepts:
\begin{equation}
\hspace{-8mm}  {\textrm {\em Cos}}_{\textrm {\em cp}}(c_1,c_2) = 
\frac{\sum_{w \in C(c_1) \cup C(c_2)} \left( P(w|c_1) \times P(w|c_2) \right) }
  {\sqrt{ \sum_{w \in C(c_1)} P(w|c_1)^2 } \times \sqrt{ \sum_{w \in C(c_2)} P(w|c_2)^2 } }
\end{equation}
\noindent $C(x)$ is now the set of words that co-occur with {\em concept} $x$
within a pre-determined window. The conditional probabilities in the formula
are taken from the distributional profiles of concepts.

\subsubsection{A suitable knowledge source and concept inventory}

\hspace{0.0pt} \namecite{MohammadH06b} use the categories in the {\it Macquarie Thesaurus}, 812 in all, as very coarse-grained
word senses or concepts, 
in contrast to approaches that use WordNet or other similarly fine-grained sense inventories.\footnote{It has been suggested for some time now
that WordNet is much too fine-grained for certain natural language applications (\namecite{AgirreL03} and citations therein).}
Their approach to determining word--concept co-occurrence counts (described in the next sub-section) requires
a set of possibly ambiguous words that together unambiguously represent each concept---for which a thesaurus is a natural choice.
The use of categories in a thesaurus as concepts means that this approach requires a concept--concept 
distance matrix of size only $812 \times 812$---much
smaller than (less than 0.01\% of) the matrix required by the WordNet-based and distributional measures.

 \subsubsection{Estimating distributional profiles of concepts}
 \label{s:DPC}

A {\bf word--category co-occurrence matrix (WCCM)}
is created having word types $w$ as one
dimension and thesaurus categories $c$ as
another.
 \begin{displaymath}
 \begin{array}{c|c c c c c}
                          &c_1      &c_2     &\ldots &c_j &\ldots\\
         \hline
 w_1     &m_{11}     &m_{12}     &\ldots     &m_{1j} &\ldots\\
 w_2     &m_{21}     &m_{22}     &\ldots     &m_{2j} &\ldots\\
 \vdots                  &\vdots     &\vdots     &\ddots         &\vdots &\vdots\\
 w_i     &m_{i1}     &m_{i2}     &\ldots     &m_{ij} &\ldots\\
 \vdots                  &\vdots     &\vdots     &\ldots     &\vdots &\ddots\\
\end{array}
\end{displaymath}
\noindent The matrix is populated with co-occurrence counts from a
large corpus.
A particular cell $m_{ij}$, corresponding to word
$w_i$ and category or concept $c_j$, is populated
with the number of times $w_i$ co-occurs (they use a window of $\pm5$ words) with any word
that has $c_j$ as one of its senses (i.e.,
$w_i$ co-occurs with any word listed under
concept $c_j$ in the thesaurus). 
This matrix, created after a first pass of the corpus, is called
the {\bf base word--category co-occurrence matrix (base WCCM)}.
A contingency
table for any particular word $w$ and category $c$ can be easily generated
from the WCCM by collapsing cells for all other words and categories
into one and summing up their frequencies.  
\begin{displaymath}
\begin{array}{c|c c}
                       &c &\neg{c}\\
     \hline
     w        &n_{wc}               &n_{w\neg}\\
     \neg{w}  &n_{\neg{c}}            &n_{\neg\neg}
\end{array}
\end{displaymath}
\noindent A suitable statistic, such as pointwise mutual information or conditional probability, will then yield the strength of association between
the word and the category.

As the base WCCM is created from unannotated text, it will be
noisy but nonetheless
capture strong word--category co-occurrence associations reasonably accurately.
 This is because the errors in determining the true category that a word co-occurs
 with will be distributed thinly across a number of other categories.
(For more discussion of the general principle see \namecite{Resnik98}.)
A second pass of
the corpus is made and the base WCCM is used to disambiguate the words in it.
A new {\bf bootstrapped WCCM} is
created such that each cell $m_{ij}$, corresponding to word
$w_i$ and concept $c_j$, is populated
with the number of times $w_i$ co-occurs with any
word 
{\em used in sense} $c_j$.
\namecite{MohammadH06a} showed that this WCCM, created after
simple sense disambiguation, better captures word--concept co-occurrence values,
and hence strengths of association values,
than the base WCCM.

\subsubsection{Mimicking semantic relatedness and semantic similarity}

The distributional profiles created by the above methodology are relation-free.
This is because (1) all co-occurring
words (not just those that are related to the target by certain
syntactic relations) are used, and (2) the WCCM, as described
above, does not maintain separate counts
for the different syntactic relations between the target and
co-occurring words. Thus, distributional measures that use
these WCCMs will estimate semantic {\it relatedness} between concepts. Distributional
measures that mimic semantic {\it similarity}, which require
relation-constrained DPCs, can easily be created from
WCCMs that have rows for each word--syntactic relation pair (rather than just for words).

\subsubsection{Performance}

\hspace{0.0pt} \namecite{MohammadH06b} evaluate this approach on two tasks: ranking word pairs in order of their semantic distance
and correcting real-word spelling errors.
On both tasks, distributional concept-distance measures markedly outperformed distributional
word-distance measures.
The WordNet-based measures performed better in the word-pair ranking task, but in the spelling correction task
three of the four distributional measures outperformed all WordNet-based measures except
the Jiang--Conrath measure. It should be noted, however, that these experiments evaluated only 
semantic similarity of noun--noun pairs---for all other part-of-speech combinations
and semantic relatedness estimates, the WordNet-based measures are markedly less accurate.

\subsection{Multilinguality}
\label{s:cross}

Some of the best algorithms for semantic distance cannot be applied to most languages
because of a lack of high-quality linguistic resources.
\namecite{MohammadGHZ07} showed how text in one language $L_1$ can be combined with a knowledge source in another $L_2$ using a bilingual lexicon $L_1$--$L_2$
and a bootstrapping and concept-disambiguation algorithm to create {\bf cross-lingual
distributional profiles of concepts}. 
These cross-lingual DPCs model co-occurrence distributions of concepts, as per
a knowledge source in one language, with words from another language,
and obtain semantic distance in a resource-poor language
using a knowledge source from a resource-rich one. 
Cross-lingual semantic distance and cross-lingual DPCs are also useful in tasks that inherently involve two or more languages,
such as machine translation, multilingual multidocument tasks of clustering, coreference resolution, and information retrieval.
We summarize their approach here using German as $L_1$ and English as $L_2$; however,
the algorithm is language-pair independent.

 \subsubsection{Cross-lingual senses, cross-lingual distributional profiles, and
 cross-lingual distributional distance}
 \label{s:crossall}

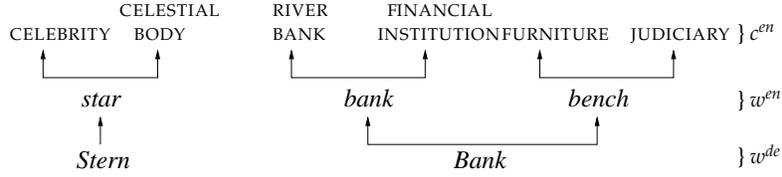
\begin{figure}[t]
\centerline{\scalebox{0.4}{ \input{stern.pstex_t} }}
\caption{The cross-lingual candidate senses of German words {\em Stern} and {\em Bank}.}
\label{fig:crossstern}
\end{figure}

Given a German word $w^{\text {\em de}}$ in context, \namecite{MohammadGHZ07} use a
German--English bilingual lexicon to determine its different possible
English translations. Each English translation $w^{\text {\em en}}$
may have one or more possible coarse senses, as listed in an English
thesaurus. These English thesaurus concepts ($c^{\text {\em en}}$)
will be referred to as the {\bf cross-lingual candidate senses} of the
German word $w^{\text {\em de}}$. 
Figure \ref{fig:crossstern} depicts examples.\footnote{They are called ``candidate senses"
  because some of the senses of $w^{\text {\em en}}$ might not really be senses
  of $w^{\text {\em de}}$. For example, {\sc celestial body} and {\sc celebrity} 
  are both senses of the English word {\it star}, but 
  the German word {\it Stern} can only mean {\sc celestial body} and 
  not {\sc celebrity}.
  Similarly, the German {\it Bank} can mean {\sc financial institution}
  or {\sc furniture}, but not {\sc river bank} or
  {\sc judiciary}. 
  An automated system has no straightforward method of teasing out the
  actual cross-lingual senses of $w^{\text {\em de}}$ from those that are an artifact of the translation step.}

As in the monolingual estimation of distributional concept-distance, the distance between
two concepts is calculated by first determining their DPs. 
However, a concept is now glossed by
near-synonymous words in an {\em English} thesaurus, whereas its
profile is made up of the strengths of association with co-occurring
{\em German} words.  
Here are constructed example cross-lingual distributional profiles of the two cross-lingual candidate senses of the German word {\em Stern}:\footnote{Vocabulary of German words needed to
understand this discussion:  
{\bf \em Bank}: 1.\@ financial institution, 2.\@ bench (furniture);
{\bf \em ber\"{u}hmt}: famous;
{\bf \em Film}: movie (motion picture);
{\bf \em Him\-mels\-k\"{o}rp\-er}: heavenly body;
{\bf \em Konstellation}: constellation;
{\bf \em Licht}: light;
{\bf \em Morgensonne}: morning sun;
{\bf \em Raum}: space;
{\bf \em reich}: rich;
{\bf \em Sonne}: sun;
{\bf \em Star}: star (celebrity);
{\bf \em Stern}: star (celestial body)
}
\begin{quote}
  {\bf \sc celestial body} ({\em celestial body, sun, \ldots}): {\em Raum} 0.36, {\em Licht} 0.27, {\it Konstellation} 0.11,~\ldots \\
  {\bf \sc celebrity} ({\em celebrity, hero,~\ldots}): {\em ber\"{u}hmt} 0.24, {\em Film} 0.14, {\em reich} 0.14,~\ldots
\end{quote}
\noindent The cross-lingual DPCs are created from a cross-lingual word--category co-occurrence matrix
without the use of any word-aligned parallel corpora or sense-annotated data (as described in the next subsection).
Just as in the case of monolingual distributional concept-distance measures, 
distributional measures can be used to estimate the distance between the cross-lingual DPs of two target concepts.
For example, the cosine formula can be adapted to estimate cross-lingual distributional distance between two concepts
as shown below:
\begin{equation}
\hspace{-8mm}  {\textrm {\em Cos}}(c_1^{\text {\em en}},c_2^{\text {\em en}}) = 
\frac{\sum_{w^{\text {\em de}} \in C(c_1^{\text {\em en}}) \cup C(c_2^{\text {\em en}})} \left( P(w^{\text {\em de}}|c_1^{\text {\em en}}) \times P(w^{\text {\em de}}|c_2^{\text {\em en}}) \right) }
  {\sqrt{ \sum_{w^{\text {\em de}} \in C(c_1^{\text {\em en}})} P(w^{\text {\em de}}|c_1^{\text {\em en}})^2 } \times \sqrt{ \sum_{w^{\text {\em de}} \in C(c_2^{\text {\em en}})} P(w^{\text {\em de}}|c_2^{\text {\em en}})^2 } }
\end{equation}
\noindent $C(x)$ is now the set of German words that co-occur with English concept $x$
within a pre-determined window. The conditional probabilities in the formula
are taken from the cross-lingual DPCs. 

\subsubsection{Creating cross-lingual word--category co-occurrence matrix}
\begin{figure}[t]
\centerline{\scalebox{0.45}{ \input{celestial.pstex_t} }}
\caption{Words having {\sc celestial body} as one of their cross-lingual
  candidate senses.}
\label{fig:crosscelestial}
\end{figure}
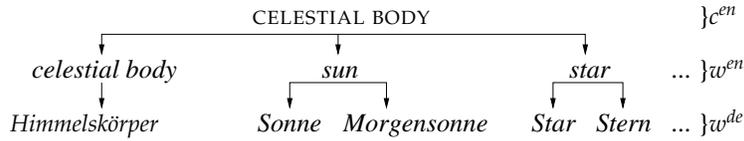

A German--English cross-lingual word--category co-occurrence matrix has
German word types $w^{\text {\em de}}$ as one dimension and English
thesaurus concepts $c^{\text {\em en}}$ as another.
 \begin{displaymath}
 \begin{array}{c|c c c c c}
                          &c_1^{\text{\em en}}      &c_2^{\text{\em en}}     &\ldots &c_j^{\text{\em en}} &\ldots\\
         \hline
 w_1^{\text{\em de}}     &m_{11}     &m_{12}     &\ldots     &m_{1j} &\ldots\\
 w_2^{\text{\em de}}     &m_{21}     &m_{22}     &\ldots     &m_{2j} &\ldots\\
 \vdots                  &\vdots     &\vdots     &\ddots     &\vdots &\vdots\\
 w_i^{\text{\em de}}     &m_{i1}     &m_{i2}     &\ldots     &m_{ij} &\ldots\\
 \vdots                  &\vdots     &\vdots     &\ldots     &\vdots &\ddots\\
\end{array}
\end{displaymath}
\noindent The matrix is populated with co-occurrence counts from a
large German corpus.
A particular cell $m_{ij}$, corresponding to word
$w_i^{\text{\em de}}$ and concept $c_j^{\text{\em en}}$, is populated
with the number of times the German word $w_i^{\text{\em de}}$
co-occurs (say a window of $\pm5$ words) with any German word having $c_j^{\text{\em en}}$ as one
of its {\em cross-lingual candidate senses}.
For example, the {\em Raum}--{\sc celestial body} cell will have the sum
of the number of times {\em Raum} co-occurs with {\em
  Himmelsk\"{o}rper, Sonne, Morgensonne, Star, Stern}, and so on (see
Figure \ref{fig:crosscelestial}).
This matrix, created after a first pass of the corpus, is called
the {\bf cross-lingual base WCCM}.
A contingency
table for any particular German word $w^{\text{\em de}}$ and English category $c^{\text{\em en}}$ can be easily generated
from the WCCM by collapsing cells for all other words and categories
into one and summing up their frequencies.  
A suitable statistic, such as PMI or conditional probability, will yield the strength of association between
the German word and the English category.
Then a new bootstrapped cross-lingual WCCM is created, just as in the monolingual case.

\subsubsection{Performance}

\hspace{0.0pt}  \namecite{MohammadGHZ07} evaluated the cross-lingual measures of
semantic distance on two tasks: (1) estimating
semantic distance between words and ranking the word pairs according
to semantic distance, and (2) solving {\em Reader's Digest} `Word
Power' problems.
They compared these results with those obtained by conventional
 state-of-the-art monolingual approaches with and without a knowledge
 source in the target language $L_1$ (GermaNet).  
 The cross-lingual approach obtained much better results
 than monolingual approaches that do not use a knowledge source.
 Further, in both tasks, the cross-lingual
 measures performed as well if not slightly better than the
 GermaNet-based monolingual approaches, as well.  This shows that the
 cross-lingual approach is able to keep losses due to the translation
 step at a minimum, while allowing the use of a superior knowledge
 source in another language to get better results.

\subsection{Challenges}
Distributional measures of concept-distance have many desirable features of both
knowledge-rich approaches and strictly corpus-based approaches---they have the high
accuracies of knowledge-rich approaches, they can measure both semantic relatedness and semantic similarity,
and they have a strong corpus-reliance making them domain adaptable. Further, with the cross-lingual approach,
a lack of high-quality knowledge source in the target language is no longer  a problem.
However, certain issues remain.

\subsubsection{Integrating domain-specific terminology}
 
The reliance on a knowledge source means that the approach cannot
measure the distance between words not listed in the thesaurus. This is especially a problem
for domain-specific jargon, which might not find place in a general purpose knowledge source.
Automatic ways of integrating domain-specific terminology into a general purpose knowledge source
will be valuable to this end.

\subsubsection{Choosing the right concept-granularity}

\hspace{0.0pt} \namecite{MohammadH06b} and \namecite{MohammadGHZ07} have reported results
using the categories of the thesaurus as very coarse word senses. This level of granularity
has worked well for the tasks they experimented with; however, a relatively finer 
sense inventory may be more suitable for other tasks. Words within a thesaurus category are grouped into paragraphs;
and using them (instead of categories) and determining when this finer-grained sense-inventory
is more suitable for use remains to be explored.

\subsubsection{Identifying lexical semantic relations}

Word pairs can be semantically close because of any of the classical lexical semantic
relations, such as hypernymy, near-synonymy, antonymy, troponymy, and meronymy, or the innumerable
non-classical relations. The various distributional approaches discussed in this paper 
determine semantic distance
without explicitly identifying the nature of the relationship. 
Already, there is some work on determining lexical entailment \cite{MirkinDG07} and determining near-synonymy \cite{LinZQZ03}.
Identifying antonymy (or more generally,
contrasting word-pairs) is especially useful in many natural language tasks, even if it is simply
to discard them from a list of distributionally close terms.
Also, it will be interesting for measures of semantic distance to 
characterize the nature of any non-classical relationship shared by two words---not only
determining if two terms are close but also specifying (in some intuitive way) the aspect of meaning they share.

\section{Conclusion}
A large number of important natural language problems, including machine translation, information retrieval,
and word sense disambiguation, can be viewed in part as semantic distance problems. 
Numerous measures of semantic distance exist---those that use a knowledge source and those that rely on corpora. Yet, their use in 
real-world applications has been limited. In this paper, we investigated how automatic measures 
can be brought more in line with human notions of semantic distance, how they can be made applicable to a large number
of natural language tasks, and how they can be used even for languages deficient in high-quality linguistic resources.

Even though corpus-based distributional measures of distance have traditionally performed poorly when compared to WordNet-based measures,
we have shown that (1) there are a number of reasons that make distributional measures uniquely attractive, and (2) that their potential
is yet to be fully realized. 
Distributional measures can be easily applied to most languages (they can make do even with just raw text) and they 
can be used to mimic both semantic similarity and semantic relatedness. 
With this in mind, the paper presented a detailed study of many important distributional measures, analyzed their limitations, and explained
why their performance has been relatively poor so far. Understanding these limitations is crucial in the development of
new and better approaches, whether they have a distributional base or otherwise.
We concluded the paper with the discussion of a hybrid, yet distinctly distributional approach, that presents
one way to more accurately measure distributional distance without compromising too much on essential
properties such as the applicability to resource-poor languages.

\begin{acknowledgments}
We thank Suzanne Stevenson, Gerald Penn, and the Computational Linguistics group at the University of Toronto for helpful discussions.
  This research is financially
   supported by the Natural Sciences and
    Engineering Research Council of Canada and the University of Toronto.
\end{acknowledgments}

\begin{multicols}{2}
\bibliography{References}
\end{multicols}

\end{document}

%% file: stern.pstex_t
\begin{picture}(0,0)%
\includegraphics{stern.pstex}%
\end{picture}%
\setlength{\unitlength}{3947sp}%
\begingroup\makeatletter\ifx\SetFigFont\undefined%
\gdef\SetFigFont#1#2#3#4#5{%
  \reset@font\fontsize{#1}{#2pt}%
  \fontfamily{#3}\fontseries{#4}\fontshape{#5}%
  \selectfont}%
\fi\endgroup%
\begin{picture}(11655,2727)(961,-4684)
\put(976,-2611){\makebox(0,0)[lb]{\smash{{\SetFigFont{20}{24.0}{\rmdefault}{\mddefault}{\updefault}{\sc celebrity}%
}}}}
\put(2926,-2611){\makebox(0,0)[lb]{\smash{{\SetFigFont{20}{24.0}{\rmdefault}{\mddefault}{\updefault}{\sc body}%
}}}}
\put(6751,-2611){\makebox(0,0)[lb]{\smash{{\SetFigFont{20}{24.0}{\rmdefault}{\mddefault}{\updefault}{\sc institution}%
}}}}
\put(2701,-2236){\makebox(0,0)[lb]{\smash{{\SetFigFont{20}{24.0}{\rmdefault}{\mddefault}{\updefault}{\sc celestial}%
}}}}
\put(5101,-2611){\makebox(0,0)[lb]{\smash{{\SetFigFont{20}{24.0}{\rmdefault}{\mddefault}{\updefault}{\sc bank}%
}}}}
\put(5101,-2236){\makebox(0,0)[lb]{\smash{{\SetFigFont{20}{24.0}{\rmdefault}{\mddefault}{\updefault}{\sc river}%
}}}}
\put(8701,-2611){\makebox(0,0)[lb]{\smash{{\SetFigFont{20}{24.0}{\rmdefault}{\mddefault}{\updefault}{\sc furniture}%
}}}}
\put(10726,-2611){\makebox(0,0)[lb]{\smash{{\SetFigFont{20}{24.0}{\rmdefault}{\mddefault}{\updefault}{\sc judiciary}%
}}}}
\put(6901,-2236){\makebox(0,0)[lb]{\smash{{\SetFigFont{20}{24.0}{\rmdefault}{\mddefault}{\updefault}{\sc financial}%
}}}}
\put(12601,-3661){\makebox(0,0)[lb]{\smash{{\SetFigFont{20}{24.0}{\rmdefault}{\mddefault}{\updefault}$w^{\text{\em en}}$%
}}}}
\put(12601,-4561){\makebox(0,0)[lb]{\smash{{\SetFigFont{20}{24.0}{\rmdefault}{\mddefault}{\updefault}$w^{\text{\em de}}$%
}}}}
\put(12601,-2611){\makebox(0,0)[lb]{\smash{{\SetFigFont{20}{24.0}{\rmdefault}{\mddefault}{\updefault}$c^{\text{\em en}}$%
}}}}
\end{picture}%

%% file: celestial.pstex_t
\begin{picture}(0,0)%
\includegraphics{celestial.pstex}%
\end{picture}%
\setlength{\unitlength}{3947sp}%
\begingroup\makeatletter\ifx\SetFigFont\undefined%
\gdef\SetFigFont#1#2#3#4#5{%
  \reset@font\fontsize{#1}{#2pt}%
  \fontfamily{#3}\fontseries{#4}\fontshape{#5}%
  \selectfont}%
\fi\endgroup%
\begin{picture}(9783,1926)(61,-5734)
\put( 76,-5611){\makebox(0,0)[lb]{\smash{{\SetFigFont{20}{24.0}{\rmdefault}{\mddefault}{\itdefault}Himmelsk\"{o}rper%
}}}}
\put(9826,-4861){\makebox(0,0)[lb]{\smash{{\SetFigFont{20}{24.0}{\rmdefault}{\mddefault}{\updefault}$w^{\text{\em en}}$%
}}}}
\put(9826,-5611){\makebox(0,0)[lb]{\smash{{\SetFigFont{20}{24.0}{\rmdefault}{\mddefault}{\updefault}$w^{\text{\em de}}$%
}}}}
\put(3451,-4111){\makebox(0,0)[lb]{\smash{{\SetFigFont{20}{24.0}{\rmdefault}{\mddefault}{\updefault}{\sc celestial body}%
}}}}
\put(9826,-4111){\makebox(0,0)[lb]{\smash{{\SetFigFont{20}{24.0}{\rmdefault}{\mddefault}{\updefault}$c^{\text{\em en}}$%
}}}}
\end{picture}%